\documentclass[letterpaper, 10 pt, conference]{ieeeconf}  %

\IEEEoverridecommandlockouts                              %

\usepackage{amsmath,amsfonts,bm,amssymb}
\usepackage{siunitx}
\usepackage{adjustbox}
\usepackage{graphicx}
\usepackage{xspace}
\usepackage{booktabs}       %
\usepackage{amsfonts}   
\usepackage{algorithm}
\usepackage[para,online,flushleft]{threeparttable}
\usepackage{subcaption}
\usepackage[noend]{algpseudocode}

\DeclareMathAlphabet{\mathcal}{OMS}{cmsy}{m}{n}

\newcommand{\algname}{{PACOH-RL}\xspace}

\def\eqref#1{equation~\ref{#1}}

\def\1{\bm{1}}

\def\E{{\mathbb{E}}}
\def\R{{\mathbb{R}}}

\DeclareMathAlphabet{\mathsfit}{\encodingdefault}{\sfdefault}{m}{sl}
\SetMathAlphabet{\mathsfit}{bold}{\encodingdefault}{\sfdefault}{bx}{n}

\DeclareMathOperator*{\argmax}{arg\,max}

\newcommand{\calB}{\mathcal{B}}

\newcommand{\calD}{\mathcal{D}}

\newcommand{\calL}{\mathcal{L}}
\newcommand{\calM}{\mathcal{M}}
\newcommand{\calN}{\mathcal{N}}

\newcommand{\calP}{\mathcal{P}}
\newcommand{\calQ}{\mathcal{Q}}

\newcommand{\calS}{\mathcal{S}}
\newcommand{\calT}{\mathcal{T}}

\newcommand{\calX}{\mathcal{X}}
\newcommand{\calY}{\mathcal{Y}}

\usepackage{ifthen}
\usepackage[normalem]{ulem}
\usepackage{xcolor}
\usepackage{etoolbox}

\definecolor{lightgrey}{RGB}{200,200,200}

\newboolean{showRevision}
\setboolean{showRevision}{false} %

\newcommand{\added}[1]{\ifthenelse{\boolean{showRevision}}{\textcolor{blue}{#1}}{#1}}

\newcommand{\edited}[2]{\ifthenelse{\boolean{showRevision}}{\textcolor{lightgrey}{\sout{#1}} \textcolor{blue}{#2}}{#2}}

\newcommand{\deleted}[1]{\ifthenelse{\boolean{showRevision}}{\textcolor{blue}{\sout{#1}}}{}}
\usepackage{hyperref}
\usepackage{cleveref}
\usepackage{xcolor}
\usepackage{tabularx}
\usepackage{graphics} %
\usepackage{epsfig} %
\usepackage{mathptmx} %
\usepackage{times} %
\usepackage{amsmath} %
\usepackage{amssymb}  %
\usepackage{gensymb}
\usepackage{float}
\usepackage{multicol}
\usepackage{multirow}
\usepackage{float}
\usepackage{dblfloatfix}

\makeatletter
\let\NAT@parse\undefined
\makeatother
\usepackage[compress, numbers]{natbib}

\usepackage{ifthen}

\newboolean{ArXivSubmission}
\setboolean{ArXivSubmission}{True}
\newcommand{\appendixref}[1]{%
    \ifthenelse{\boolean{ArXivSubmission}}{#1}{}%
}

\title{\LARGE \bf
Data-Efficient Task Generalization via \\Probabilistic Model-based Meta Reinforcement Learning}

\author{Arjun Bhardwaj$^{1}$, Jonas Rothfuss$^{1}$, Bhavya Sukhija$^{1}$, Yarden As$^{1}$, \\Marco Hutter$^{1}$, Stelian Coros$^{1}$, Andreas Krause$^{1}$
 \thanks{*This work was not supported by any organization}%
 \thanks{$^{1}$ETH Zurich, correspondence to \texttt{abhardwaj@ethz.ch}}
}

\begin{document}

\maketitle
\thispagestyle{empty}
\pagestyle{empty}

\begin{abstract}

\looseness -1 We introduce PACOH-RL, a novel model-based Meta-Reinforcement Learning (Meta-RL) algorithm designed to efficiently adapt control policies to changing dynamics. %
PACOH-RL meta-learns priors for the dynamics model, allowing swift adaptation to new dynamics with minimal interaction data.
Existing Meta-RL methods require abundant meta-learning data, limiting their applicability in settings such as robotics, where data is costly to obtain. To address this, PACOH-RL incorporates regularization and epistemic uncertainty quantification in both the meta-learning and task adaptation stages. When facing new dynamics, we use these uncertainty estimates to effectively guide exploration and data collection.
Overall, this enables positive transfer, even when access to data from prior tasks or dynamic settings is severely limited. 
Our experiment results demonstrate that PACOH-RL outperforms model-based RL and model-based Meta-RL baselines in adapting to new dynamic conditions. Finally, on a real robotic car, we showcase the potential for efficient RL policy adaptation in diverse, data-scarce conditions.

\end{abstract}
\section{Introduction}
\label{sec:intro}

\looseness -1 The field of Reinforcement Learning (RL) has seen remarkable advances in recent years \added{\citep[e.g.,][]{mnih2015human, rajeswaran2017learning}}. Particularly, in gameplay and simulated robotic manipulation problems, RL agents \edited{are able to}{can} solve ever more complex tasks. These advances, however, largely rely on an abundance of agent-environment interactions. In contrast, in real-world robotic applications, obtaining interaction data is costly. Thus, a promising approach for real robotic platforms is {\em Model-based Reinforcement Learning (MBRL)}, which leverages a learned model of the environment dynamics to obtain a control policy more efficiently \added{\citep[e.g.,][]{deisenroth2011pilco, chua2018deep, clavera2018model}}.

Still, \added{this requires} a significant amount of data to learn a reliable dynamics model. This becomes even more problematic when generalization across multiple tasks with related but different dynamics is required,
e.g., when we want to deploy the same robot on different surfaces or with different payloads/tools. 
The challenges above give rise to the following question:
{\em How can we effectively transfer knowledge across tasks and adapt our model to changes in dynamics without the need for extensive data collection each time?}

\looseness -1 Addressing this challenge, we propose \textsc{PACOH-RL}, a novel approach to model-based Meta-Reinforcement Learning (Meta-RL) which allows us to transfer prior experience on the robotic platform to new dynamics conditions. In particular, we propose to {\em meta-learn a prior distribution over the dynamics model}, facilitating efficient (Bayesian) adaptation to new settings based on minimal interaction data. 
Unlike existing Meta-RL methods \added{\citep[e.g.,][]{duan2016rl, nagabandi2018}}, our approach takes into account epistemic uncertainty, both throughout the meta-learning and task adaptation stages.
This \edited{enables}{allows} us to use an RL formulation that plans optimistically w.r.t. the epistemic uncertainty of \added{the dynamics model \citep[see][]{curi2020efficient}}. As a result, \textsc{PACOH-RL} explores uncertain actions that plausibly lead to high rewards in a directed manner and, thus, efficiently adapts the model to the current dynamics conditions.

\begin{figure}[t]
\vspace{6pt}
    \centering
    \includegraphics[width=0.48\textwidth]{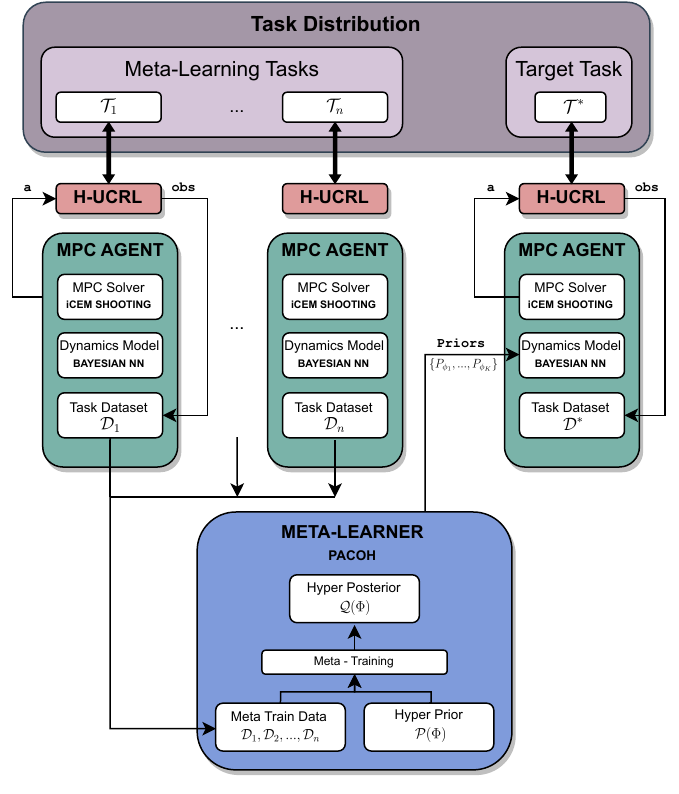}
    \caption{The \algname framework uses datasets of transitions $\calD_1, ..., \calD_n$ from previous RL tasks to meta-learn a BNN prior. Then, we equip our BNN dynamics model with the meta-learned prior. This significantly improves the sample efficiency of model-based RL on a new target task.}
    \label{fig:pacoh_framework}
\vspace{-18pt}
\end{figure}

Crucially, our method is designed to operate {\em with only a handful of previous tasks} (i.e., dynamics settings), a regime where existing Meta-RL methods typically fail.
Such efficiency is essential in real-world applications where collecting data across different tasks with different dynamics is limited. 

\looseness -1 Our experiments show that \textsc{PACOH-RL} can quickly adapt to new dynamics and consistently outperforms both model-based Meta-RL and standard model-based RL baselines. 
Thanks to the combination of uncertainty-aware meta-learning and directed exploration, \textsc{PACOH-RL} performs particularly favorably in RL environments with sparse rewards. 
Finally, we showcase how PACOH-RL facilitates successful transfer on a real robotic car with only a handful of meta-training datasets. This demonstrates the practicality of our approach and highlights its potential to efficiently adapt RL policies to changing conditions.

\section{Related Work}
\label{sec:related_work}

\looseness=-1 \textbf{Model-Based Reinforcement Learning. }
\added{MBRL} methods are often considered for learning directly on hardware since they are generally more sample-efficient compared to model-free RL \cite{deisenroth2011pilco,chua2018deep}. 
However, asymptotically, model-free methods often outperform \added{MBRL}. One reason for this performance gap is the exploitation of model inaccuracies \cite{chua2018deep, clavera2018model}. Using dynamics models that are aware of epistemic uncertainty, such as ensembles or Bayesian Neural Networks (BNN), has been shown to alleviate the problem of model exploitation \citep{chua2018deep}. Building on this insight, we center our approach around BNN dynamics models. However, unlike learning a policy that is robust w.r.t. the model's uncertainty, we use the optimistic RL objective of \cite{curi2020efficient} that incentivizes exploration and data collection in areas where the model is uncertain. This allows us to reduce our model's inaccuracies more efficiently.

\looseness=-1 \textbf{Meta-Learning and Meta-RL. } 
Meta-learning \citep{Thrun1998, finn2017model, pavasovic2022mars} aims to acquire useful inductive bias from a set of related learning tasks, allowing us to \emph{quickly adapt} to a new, similar task.
\added{Some} model-free Meta-RL approaches achieve this by training a sequence model to act as reinforcement learner \citep{wang2016learning, duan2016rl}, meta-learning a policy initialization that can be quickly adapted to new tasks \cite{finn2017model, rothfuss2019promp} or conditioning the policy on a latent context variable that represents the different tasks \cite{rakelly2019efficient, Luo2022AdaptTE}.
However, model-free Meta-RL methods require \added{many} tasks during meta-training. This renders them infeasible for real-world robotic problems. 

\looseness -1 Model-based meta-RL is more task and sample efficient as it focuses on meta-learning inductive bias for the dynamics model, and does not require online control of the robot during meta-training. 
A common model-based meta-RL approach is to
condition the dynamics model on latent task variable \citep{saemundsson2018meta, perez2020generalized, lee2020context, hiraoka2021meta}. Another method, proposed by \cite{nagabandi2019learning}, is to meta-learn an NN initialization for the dynamics model using MAML \cite{finn2017model}.
Our approach builds on PAC-Bayesian meta-learning \cite{pentina2014pac, amit2018meta, rothfuss2021pacoh, rothfuss2023meta}. In particular, we employ the PACOH-NN method \cite{rothfuss2021pacoh} to meta-learn priors for our BNN dynamics models, which are adapted to the target task via (generalized) Bayesian inference. Our proposed method \added{distinguishes} itself from previous model-based meta-RL in the following two key characteristics. First, it features principled regularization and captures epistemic uncertainty both during the meta-learning and the task adaptation stage. Second, it uses the resulting uncertainty estimates to guide exploration and data collection to facilitate more efficient task adaptation. This allows PACOH-RL to perform positive transfer with only a handful of meta-learning datasets---a setting where previous approaches typically fail.

\section{Background}
\label{sec:background}
\textbf{Reinforcement Learning. } A discrete-time Markov decision process (MDP) is defined by the tuple $\calT = (\mathcal{S}, \mathcal{A}, p, p_0, r, T)$. 
Here, $\mathcal{S} \subseteq \R^{d_s}$ is the state space, $\mathcal{A} \subseteq \R^{d_a}$ the action space, $p(s_{t+1}|s_t, a_t)$ the transition distribution, $p_0$ represents the initial state distribution,
$r: \mathcal{S} \times \mathcal{A} \rightarrow \R$ is a known reward function, and $T$ the time horizon. 
For ease of notation, we exclude the discount factor $\gamma$ in the ensuing discussions. %
We then define the return $R(\tau)$ as the cumulative sum of rewards along a trajectory $\tau := (s_{0}, a_{0}, ..., s_{T-1}, a_{T-1}, s_{T})$. The central objective of reinforcement learning is to find a policy $\pi(a|s)$ that optimizes the expected return $\E_{\tau \sim P_{\mathcal{T}}(\tau | \pi)} \left[ \sum_{t=0}^{H-1} r(s_t, a_t) \right]$. Here, $P_{\mathcal{T}}(\tau | \pi) = p_0(s_0)\prod_{t=0}^H \pi(a_t| s_t) p(s_{t+1}|s_t, a_t)$ is the trajectory distribution under the MDP $\mathcal{T}$ and the policy $\pi$.

\looseness -1 \textbf{Model-based RL. }
\added{MBRL} uses the collected state transition data $\calD=\{(s_t, a_t, s_{t+1}) \}$ to learn/estimate a model $\hat{p}(s_{t+1}|s_t, a_t)$ of the transition distribution, also referred to as a {\em dynamics model}. Often, function approximators such as neural networks are employed for this purpose. Then, the estimated dynamics model is either used to simulate trajectories to train a policy or to produce dynamics predictions/constraints for a controller. Generally, the dynamics model empowers the agent to simulate future states and rewards. Through that, the agent can anticipate the outcome of its actions without interacting with its MDP environment directly. Hence, MBRL methods are typically much more data-efficient than model-free methods. 

\looseness -1 \textbf{Bayesian Neural Networks. }
Learning the dynamics model in model-based RL is a standard supervised learning problem. The inputs correspond to a concatenation of the current state and action, i.e., $x=[s, a]$, and the prediction target is the next state, i.e., $y=s'$, allowing us to write the training data set as $\calD = \left\{ (x_j, y_j) \right\}_{j=1}^m$.
Let $h_\theta: \calX \mapsto \calY$ be a function parameterized by a neural network (NN) with weights $\theta \in \Theta$. Using the NN mapping, we can define a conditional predictive distribution (a.k.a. likelihood function) $p(y|x,\theta) = \calN(y|h_\theta(x), \sigma^2)$, where $\sigma^2$ is the variance corresponding to the aleatoric uncertainty.
 
\looseness -1 Some \added{MBRL} methods (e.g., \cite{Heess2015LearningCC, nagabandi2018}) simply fit a single NN $h_\theta$ by maximizing likelihood $p(\calD | \theta) = \prod_{j=1}^m p(y_j|x_j,\theta)$ of the data. However, with a simple NN we cannot quantify {\em epistemic uncertainty}, which is crucial for directed exploration \cite{ curi2020efficient, houthooft2016curiosity}.
In contrast, {\em Bayesian Neural Networks (BNNs)} maintain a distribution over NNs, allowing them to quantify uncertainty about $h_\theta(\cdot)$. In particular, BNNs presume a prior distribution $p(\theta)$ over the model parameters $\theta$, which they combine with the data likelihood into a (generalized) posterior distribution $p(\theta | \calD) \propto p(\calD | \theta)^\lambda p(\theta)$. Note that in this paper, we resort to generalized Bayesian learning  \cite{grunwald2012safe, guedj2019primer} where the likelihood is tempered with \added{$\lambda \in (0, 1)$}, giving us additional robustness when the standard Bayesian assumptions are violated. To make probabilistic predictions, we typically form the predictive distribution as $p(y^*|x^*,\calD) = \int p(y^*|x^*,\theta) p(\theta|\calD) d\theta$ by marginalizing over the NN parameters $\theta$.

\looseness -1 \textbf{Approximate Inference via SVGD. } Since BNN posteriors are generally intractable, approximate inference techniques such as MCMC \cite{welling2011bayesian}, variational inference (VI) \cite{blei2017variational} or particle VI methods \cite{liu2016stein, chen2018unified} are often applied. In this paper, we employ {\em Stein Variational Gradient Descent (SVGD)} \cite{liu2016stein} which approximates the posterior $p(\theta | \calD)$ by a set of \added{$L$ particles $\{\theta_1, ..., \theta_L \}$}. After initialization, SVGD iteratively transports the particles (here: NN parameters) to match $p(\theta | \calD)$ by applying a form of functional gradient descent that minimizes the KL divergence in the reproducing kernel Hilbert space induced by a kernel function $k(\cdot,\cdot)$. In particular, the update of a particle $\theta$ is computed as 
\added{
\begin{equation}\label{eq:svgd_update}
\hspace{-2pt} \psi(\theta) = \frac{1}{L} \sum_{l'=1}^L \left [k(\theta_{l'}, \theta) \nabla_{\theta_{l'}} \log p(\theta_{l'} | \calD) +  \nabla_{\theta_{l'}} k(\theta_{l'}, \theta) \right] \hspace{-2pt}
\end{equation}}
and applied via \added{$\theta_l \leftarrow \theta_l + \eta \psi(\theta_l)$} where $\eta$ is the step size. While the first term in (\ref{eq:svgd_update}) moves the particles towards areas of higher probability, the second term, i.e., \added{$\nabla_{\theta_{l'}} k(\theta_{l'}, \theta)$}, acts as a repulsion force among the particles which ensures that they are well dispersed throughout the parameter space and do not collapse in the mode of the distribution.

\section{PACOH-RL: Uncertainty-Aware Model-Based Meta-RL}
\label{sec:method}

\subsection{Problem Statement: Meta-RL}
We study the problem of Meta-RL where we face multiple MDPs that vary in their transition dynamics.
While we could learn a policy that is robust to the varying dynamics, such policies are typically sub-optimal due to over-conservatism \cite{lim2013reinforcement}. Instead, we want to swiftly adapt our agent's behavior to the new dynamical conditions without requiring a large amount of agent-environment interactions.

Formally, we are given a sequence of MDP tasks $\calT_1, ..., \calT_n \sim p(\calT)$ with $\calT_i = (\mathcal{S}, \mathcal{A}, p_i, p_0, r, H)$ where the transition probabilities $p_i(s_{t+1} | s_t, a_t)$ differ between the tasks. Our framework also seamlessly supports reward functions that may vary across tasks. However, for simplicity, we treat the reward as fixed throughout the remainder of the paper.

\looseness -1 Suppose we have already collected (e.g., using RL) transition data $\calD_i = \{(s, a, s') \}$ for $n$ tasks corresponding to $p_i$, for $i=1, ..., n$. Now, we are facing a new target task $\calT^* \sim p(\calT)$ for which we want to efficiently find an optimal policy. In particular, we focus on real-world robotic settings where $n$, i.e., the number of previous tasks, is small and the agent-environment interactions are costly. Hence, we want to explore the target task and find an optimal policy for it with as few interactions as possible.
Out of this problem setting arise two key questions: 1) How can we transfer knowledge from the previously collected transition datasets $\calD_1, ... \calD_n$ to the new target task? 2) Which RL paradigm and exploration scheme should we use on the target task?

\subsection{Our approach:  Model-Based Meta-RL}
Since data efficiency is one of our core concerns, we employ a model-based RL paradigm.
To transfer knowledge from the previous MDP tasks, we use meta-learning to extract inductive bias from the transitions of previous tasks $\calD_1, ... \calD_n$, which we then harness when estimating a dynamics model on the target task. Crucially, we choose a meta-learner and dynamics model that can reason about epistemic uncertainty. When performing RL on the target task, this allows us to explore in a directed manner towards areas of the state-action space in which we are more uncertain yet can plausibly obtain high rewards. As a result, our agent is able to swiftly collect transition data on the target task, making the dynamics model more accurate, and the resulting policy better. In the following, we explain the building blocks of our approach in more detail.

\begin{algorithm*}[th]
\caption{\strut PACOH-RL (MPC version)}\label{alg:pacoh_rl}
\hspace*{\algorithmicindent} \textbf{Input:} 
Transition datasets $\{\calD_1,\ldots, \calD_n\}$ from previous tasks, test task $\mathcal{T}$, hyper-prior $\calP$
\begin{algorithmic}[1]
\State $\{P_{\phi_1}, \ldots, P_{\phi_K} \} \leftarrow \textsc{PACOH-NN}(\calD_1,\ldots, \calD_n, \calP)$ \hfill \Comment{Meta-learn set of priors to approx. $\calQ$}
\State $\calD^* \leftarrow \emptyset$  \Comment{Initialize empty transition dataset}
\For{$\textnormal{episode}=1,2,\ldots$}
    \For{$k=1,2,\ldots,K$} \label{alg_line:svgd1}
        \State $\Theta_k \leftarrow \textsc{BNN-SVGD}(\calD^*, P_{\phi_k})$ \label{alg_line:svgd2} \Comment{Train BNN with latest transition data}
    \EndFor
    \State $\Theta := \{ \Theta_1, ..., \Theta_K\}$
    \State $\hat{p}_\Theta \gets \calN(\hat{\mu}_{\Theta}(s, a), \hat{\sigma}^2_{\Theta}(s, a))$ \Comment{Aggregate NN predictions into predictive distribution}
    \State $(s_0, a_0, ..., a_{T-1}, s_T) \leftarrow \textsc{iCEM-MPC}(\hat{p}_\Theta, \calT)$ \Comment{Perform rollout with MPC controller} \label{alg_line:mpc_rollout}
    \State $\calD^* \leftarrow \calD^* \cup \{(s_t, a_t, s_{t+1})\}_{t=0}^{T-1}$  \Comment{Add transitions to dataset}
\EndFor
\end{algorithmic}
\end{algorithm*}

\textbf{Meta-Learning a dynamics model prior. }
Due to its principled meta-level regularization, which allows successful meta-learning from only a handful of tasks as well as its principled treatment of uncertainty, we build on the {\em PAC-Bayesian meta-learning} framework \cite{pentina2014pac, amit2018meta, rothfuss2021pacoh}. In particular, we employ \textsc{PACOH-NN} \cite{rothfuss2021pacoh} which meta-learns Bayesian Neural Network (BNN) priors from the meta-training data $\calD_1, ..., \calD_n$.
The \textsc{PACOH} framework uses a parametric family of priors $\{ P_\phi | \phi \in \Phi\}$ over NN parameters $\theta$. Due to computational convenience, we use Gaussian priors, i.e., \added{$P_{\phi_k} = \calN(\mu_{P_k}, \text{diag}(\sigma_{P_k}^2))$ with $\phi_k:= (\mu_{P_k}, \ln \sigma_{P_k})$}. The prior variance $\sigma_{P_k}^2$ is represented in the log-space to avoid additional positivity constraints. Employing a hyper-prior $\mathcal{P}(\phi)$ which acts as a regularizer on the meta-level, the meta-learner infers the hyper-posterior, a distribution over the prior parameters, in particular, 
\begin{equation}
\label{eq:pacoh}
    \mathcal{Q}(\phi) \propto \mathcal{P}(\phi) \exp\left (\frac{1}{\sqrt{nm}+1}\sum_{i=1}^{n} \ln Z(\phi, \calD_i)\right ) ~.
\end{equation}
\looseness -1 Here $\ln Z(\phi, \calD_i) = \ln E_{\theta \sim P_\phi} \left[ p(\calD | \theta)^{1/\sqrt{n}} \right]$ denotes the generalized marginal log likelihood (MLL). Sampling from and determining the normalization constant of $\mathcal{Q}(\phi)$ is challenging. Thus, we follow \cite{rothfuss2021pacoh} and approximate $\mathcal{Q}(\phi)$ by a set of \added{$K$ priors $P_{\phi_1}, ..., P_{\phi_K}$} which are optimized via Stein Variational Gradient Descent (SVGD) \cite{liu2016stein} to closely resemble $Q(\phi)$. By considering a distribution over priors rather than meta-learning a single prior, PACOH is able to quantify epistemic uncertainty on the meta-level. When the number of meta-learning tasks, i.e. $n$, is small, the sum of generalized \added{MLLs} in (\ref{eq:pacoh}) is relatively small, and the hyper-prior keeps the uncertainty in $\calQ$ large. As we have more meta-learning tasks, the exponential term in (\ref{eq:pacoh}) grows, and $\calQ$ becomes increasingly peaked in prior parameters that yield a large MLL across the tasks, reflecting reduced uncertainty on the meta-level.

\looseness -1 \textbf{Adapting the dynamics model to the target task. }
From the meta-learning stage, we have acquired a set of priors $P_{\phi_1}, ..., P_{\phi_K}$ from the transitions of previous tasks, which give us good inductive bias towards the dynamics of our robot under varying conditions. Once we observe state transitions under the dynamical conditions of the target task $\calT^*$, we can combine these empirical observations with our meta-learned prior knowledge into a BNN model. Let $\calD^*$ be the dataset of observed transitions on the target task. 
Then, the generalized BNN posterior corresponding to the meta-learned prior $P_{\phi_k}$ follows as $Q_k(\theta; \calD^*) \propto p(\calD^*|\theta)^{1/\sqrt{n}} P_{\phi_k}(\theta)$. 
Since we have $K$ priors, we also obtain $K$ different posteriors, i.e., $Q_k(\theta; \calD^*) ~ k=1, ..., K$. Similar to the meta-learning stage, we represent each posterior $Q_k$ by a set of $L$ \added{NN} parameters $\Theta_k=\{\theta_{k,1}, ..., \theta_{k,L} \}$ which we optimize via SVGD to approximate the posterior density. 
This leaves us with $K \cdot L$ neural networks whose parameters we denote by \added{$\Theta = \{ \Theta_1 ,..., \Theta_K\}$}. \added{The $K$ sets of NN particles represent the epistemic uncertainty on the meta-level whereas the particles within $\Theta_k$ correspond to the uncertainty on the target task. In all experiments, we use $K=L=3$.}
To aggregate the individual neural networks' predictions into a predictive distribution, we use a Gaussian approximation $\hat{p}_{\Theta}(s'| s, a) = \calN(s' ; \hat{\mu}_{\Theta}(s, a), \hat{\sigma}^2_{\Theta}(s, a))$ where $\hat{\mu}_{\Theta}(s, a) = \frac{1}{KL} \sum_{k=1}^K \sum_{l=1}^L  h_{\theta_{k,l}}(s, a)$ is the predictive mean and $\hat{\sigma}^2_{\Theta}(s, a) = \frac{1}{KL} \sum_{k=1}^K \sum_{l=1}^L \left(h_{\theta_{k,l}}(s, a) - \hat{\mu}(s, a)\right)^2$ the epistemic variance.

\looseness -1 \textbf{Model-based control and exploration. } \label{sec:mb_control}
When performing model-based RL, we alternate between performing trajectory rollouts on the real environment %
and updating our dynamics model and policy with the newly collected data. This raises two important questions: {\em How to formulate and solve the model-based control/policy search problem? How to explore and collect informative transition data?}

A key feature of our BNN dynamics models is their ability to quantify epistemic uncertainty. We harness these uncertainty estimates to perform uncertainty-guided exploration. %
In particular,
we employ {\em hallucinated upper-confidence reinforcement learning (H-UCRL)} \cite{curi2020efficient} which explores by planning optimistically w.r.t.~the dynamics model's epistemic uncertainty. It hallucinates auxiliary controls $\eta(s, a) \in [-1, 1]^{d_s}$ that allow the policy to choose any state transition that is plausible within the (epistemic) confidence regions $[\hat{\mu}(s, a) \pm  \nu \hat{\sigma}(s, a)]$ of the dynamics models. This results in the following optimistic H-UCRL RL objective:

\begin{equation}\label{eq:hucrl_policy_optimization}
\begin{split}
   \pi&^*  = \argmax_\pi \max_{\eta}~ \E_{a_t \sim \pi(a_t | s_t)} \left[ \sum_{t=0}^{T-1} r(s_t, a_t) \right]  \\
   &\text{ s.t. }  s_{t+1} = \hat{\mu}(s_t, a_t) + \nu \eta(s_t, a_t) \hat{\sigma}(s_t, a_t)
\end{split}
\end{equation}

\looseness -1The objective in (\ref{eq:hucrl_policy_optimization}) steers the policy to areas of the state-action space with high reward and high epistemic uncertainty. By collecting transition data from areas prone to inaccurate predictions, the BNN dynamics becomes quickly more accurate, which facilitates efficient adaptation to the target task. 
As we collect more data, \deleted{the} the epistemic variance $\hat{\sigma}^2(s, a)$ decreases, which leads to less and less exploration, ultimately ensuring convergence to an optimal policy.

\looseness -1
To solve the control/policy-search problem in (\ref{eq:hucrl_policy_optimization}) we propose to use either a model-predictive control (MPC) solver, in particular, the improved cross-entropy method (iCEM) \cite{icem}, or use policy search on the learned dynamics model. %
MPC-based approaches often perform favorably in the context of model-based RL since they are more robust w.r.t.~model inaccuracies due to the constant re-planning \citep[cf.][]{chua2018deep}. However, compared to neural network-based policies, they are also much more computationally expensive at deployment time and tend to exhibit limitations when applied to high-dimensional action spaces. For scenarios with computational constraints or real-time requirements, e.g., on a real robot, we propose to learn an NN policy. Specifically, we employ Soft Actor-Critic (SAC) \cite{sac} with the BNN dynamics model as a swap-in for the real environment to learn a policy that optimizes the objective in (\ref{eq:hucrl_policy_optimization}). In this work, we evaluate our method with iCEM and SAC\appendixref{(see \Cref{sec:method_details})}. However, our method is flexible enough to be used alongside most ``off-the-shelf'' MDP solvers and controllers.

\label{sim_experiments}
\begin{figure*}[th]
    \centering
    \includegraphics[width=\textwidth]{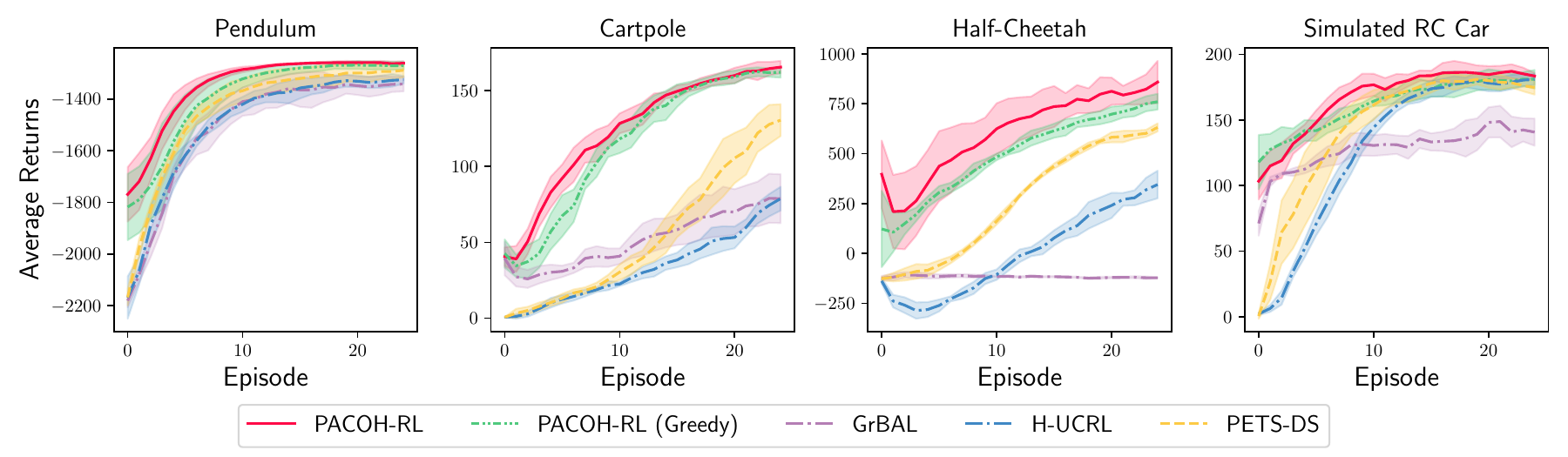}
    \caption{Returns on evaluation tasks averaged over five seeds. We compare \algname to its greedy counterpart, \algname (greedy), GrBAL \cite{nagabandi2019learning}, GrBAL-2x, H-UCRL \citep{curi2020efficient}, and PETS-DS \citep{chua2018deep}. For all the environments, \algname systematically outperforms the baselines in terms of sample efficiency and average return.}
    \label{fig:sim_normal_test}
    \vspace{-12pt}
\end{figure*}

\looseness - 1\textbf{Overview of the Approach. }
After we have introduced the building blocks, we provide an overview of our model-based meta-RL approach, which we refer to as \textsc{PACOH-RL}. The MPC version of our method is summarized in Alg. \ref{alg:pacoh_rl} and schematically illustrated in Fig. \ref{fig:pacoh_framework}\appendixref{(for the SAC version, see Algorithm \ref{alg:pacoh_rl_sac} in the appendix)}.
Given the transition data $\calD_1, \ldots, \calD_n$ from previous tasks, we form a particle approximation of the hyper-posterior $\calQ(\phi)$ with SVGD, resulting in a set of BNN priors $\{P_{\phi_1}, \ldots, P_{\phi_K}\}$. Each prior reflects our meta-learned prior knowledge about the general dynamics of our robot. The differences among the priors reflect the epistemic uncertainty due to the limited number of tasks and samples per task available for meta-learning.

Equipped with the meta-learned priors, we move on to model-based RL on the target task $\calT^*$, which we aim to solve. Since our agent has not yet interacted with $\calT^*$, we start with an empty dataset $\calD^*$ of transitions. Then, we iteratively alternate between fitting/updating our BNN dynamics models to the latest transition dataset $\calD^*$ and rolling out one episode with our control policy based on the updated dynamics model $\hat{p}_\Theta$. At the end of each episode, we add the corresponding transition tuples to $\calD^*$ and repeat the process. Initially, when $\calD^*$ is empty, our BNN dynamics model reflects the meta-learned prior, constituting a much better model starting point than classical model-based RL methods without meta-learning. With every episode, our transition dataset grows in size, allowing the model to quickly adapt to the current dynamics conditions of the target task. As the dynamics model becomes more accurate, the performance of the model-based control policy also improves.

\section{Experiments}
\label{sec:experiments}
\begin{figure}[t]
    \centering
    \includegraphics[width=0.48\textwidth]{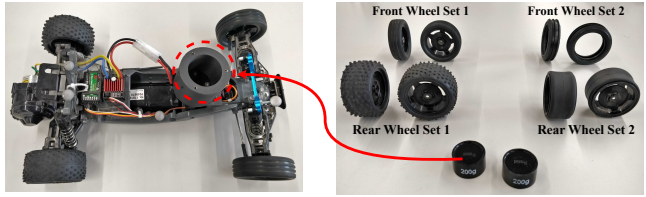}
    \caption{High torque motor RC car used in the hardware experiments. As depicted on the right, we have two different tire profiles for both the front and rear wheels. We can also add up to \SI{400}{\gram} of weight to the front of the car in a cylindrical box encircled in the image.\vspace{-10pt}}
    \label{fig:rc_car_setup}
\end{figure}

We evaluate \algname in simulation and on hardware, in particular, a highly dynamic remote-controlled (RC) race car (see~ Fig. \ref{fig:rc_car_setup}). In our experiments, we investigate the following three aspects; (\emph{i}) Does \algname improve sample efficiency on the target task?, (\emph{ii}) does the uncertainty-aware, optimistic RL formulation in (\ref{eq:hucrl_policy_optimization}) improve exploration in environments with sparse rewards, and (\emph{iii}) does \algname facilitate successful transfer on real-world hardware systems?

\begin{figure*}[th]
    \centering
    \includegraphics[width=0.96\textwidth]{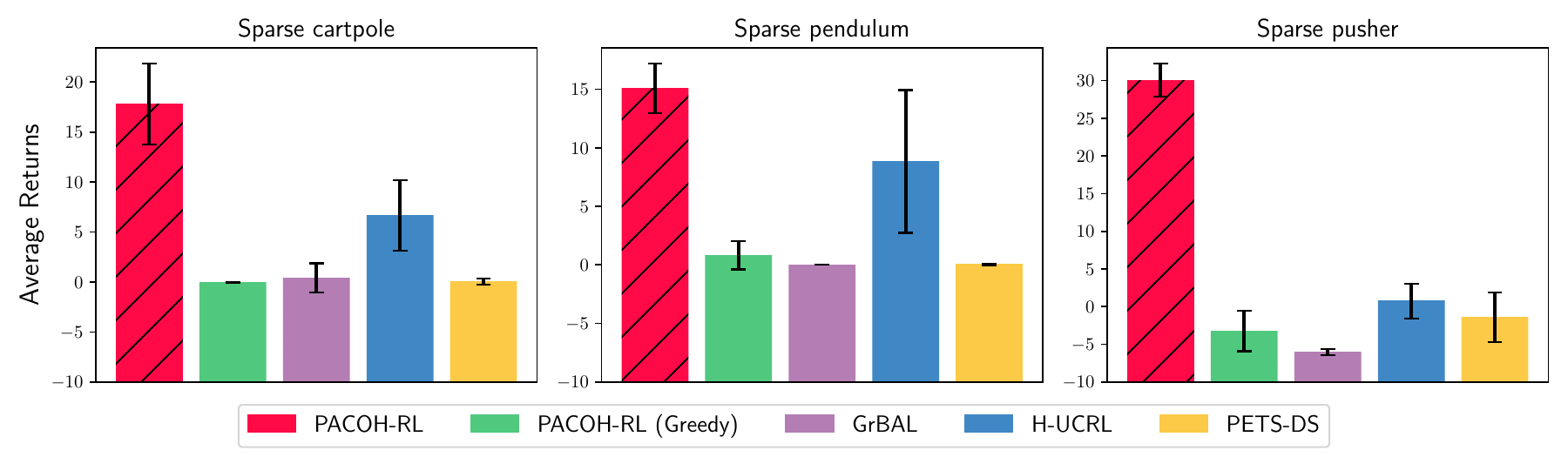}
    \caption{Returns after learning on evaluation tasks with sparse rewards for 10 episodes over five different seeds. We compare \algname to its greedy counterpart, \algname (greedy), H-UCRL, and its greedy version PETS-DS. In all environments, optimistic planning outperforms its greedy counterpart, with \algname performing the best.}
    \label{fig:sparse_rewards_exps}
    \vspace{-10pt}
\end{figure*}
\subsection{Simulation Experiments}
\label{sec:sim_experiments}
For our simulation experiments, we consider the Pendulum, Cartpole, and Half-Cheetah environments from OpenAI gym~\cite{brockman2016openai}, and a simulated model remote-controlled (RC) car. 
The RC car simulator is based on realistic race car dynamics used for autonomous racing~\cite{kabzan2020amz}.
We use a time-truncated version of the Half-Cheetah environment where the episode is terminated after 250 timesteps instead of 1000. Since the goal of PACOH-RL is to facilitate transfer and efficient adaptation to new dynamics settings, we vary the physical parameters of the simulation environment in our empirical studies.
To showcase generalization in a realistic low-task regime, we sample 20 dynamical settings/tasks at random for the meta-training and 5 for evaluation. For all our experiments, we repeat the task generation and sampling procedure with five seeds and report the mean estimate along with the standard deviation of the achieved returns averaged over the 5 evaluation tasks. \appendixref{We provide more details on the environments and experimental setup in Fig.~\ref{sec:details_exps}.}

\textbf{Does \algname improve the efficiency of RL on the target task? }
To demonstrate the sample efficiency of \algname, 
we compare it to two model-based RL algorithms; H-UCRL~\cite{curi2020efficient} and PETS with distribution sampling (PETS-DS~\cite{chua2018deep}). Furthermore, we also compare \algname to the gradient-based adaptive learner (GrBAL) algorithm \cite{nagabandi2019learning}, a state-of-the-art model-based meta-RL method that is based on MAML~\cite{finn2017model} for meta-training. As an ablation study, we also compare a greedy version of \algname, which does not use the optimistic planning objective in (\ref{eq:hucrl_policy_optimization}), and, instead, greedily maximizes the RL objective with PETS-DS while being robust w.r.t. the epistemic uncertainty in the dynamics model. We call this variant \algname (Greedy). To ensure comparability, we use the iCEM-MPC controller for all methods. 
Fig~\ref{fig:sim_normal_test} reports the average returns for all the methods over the course of 25 episodes/trajectories.

We observe that thanks to its meta-learned BNN prior, \algname starts off with a significantly better policy than the baselines. Importantly, it still is able to improve performance quickly and maintain its advantage over the other methods. Overall, \algname's ability to achieve higher rewards with fewer episodes demonstrates the effectiveness of the meta-learned BNN priors towards improving the sample efficiency of model-based RL. 
Unlike \algname, we observe that GrBAL \cite{nagabandi2019learning} often stagnates in performance early on, or only improves slowly as it collects more trajectories. We hypothesize that the observed negative transfer of GrBAL is because MAML overfits the few meta-training tasks. \appendixref{We also report results for the case where GrBAL is trained with higher number of meta-training tasks in~\Cref{fig:sim_normal_test_grbal2x}. }Despite the very limited meta-learning data, \algname is able to achieve positive transfer which we attribute to the principled meta-level regularization and treatment of epistemic uncertainty of the approach.

\textbf{Does the uncertainty-aware, optimistic RL formulation in (\ref{eq:hucrl_policy_optimization}) improve exploration in environments with sparse rewards? }
\looseness=-1
In Fig.~\ref{fig:sim_normal_test}, PACOH-RL with the optimistic exploration performs slightly better than the greedy version in the majority of environments. However, the difference between them is small because the environments have dense rewards and, thus, require little exploration. 

\looseness -1 The authors of \cite{curi2020efficient} show that in environments with sparse reward signals, principled exploration becomes much more crucial. To further investigate the influence of the optimistic planner from (\ref{eq:hucrl_policy_optimization}), we perform experiments on the Pendulum, Cartpole, and Pusher environments with sparse rewards.
We compare \algname to its greedy variant and, as non-meta-learning baselines, report the performance of H-UCRL and its greedy counterpart PETS-DS. We train all agents for 25 episodes and report their average returns over the last ten episodes in Fig.~\ref{fig:sparse_rewards_exps}.
As we can observe, the optimistic, uncertainty-aware planning objective of PACOH-RL considerably improves the agent's ability to achieve high returns in sparse reward environments. This is similarly true for both H-UCRL and \algname, and is in line with the results in \cite{curi2020efficient}.

\begin{figure}[th]
\centering
\begin{subfigure}{.49\textwidth}
  \centering
  \includegraphics[width=0.95\linewidth]{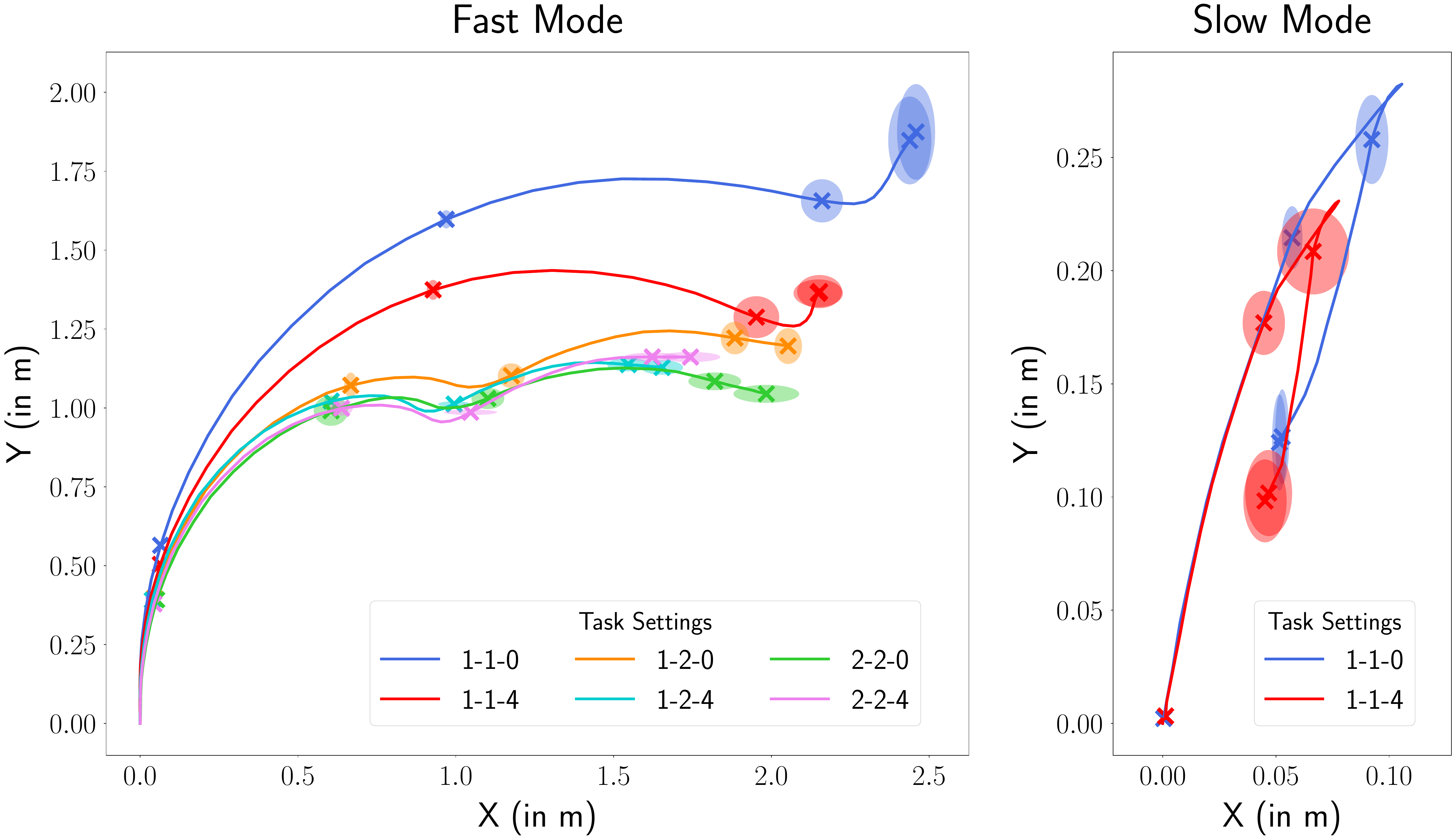}
\end{subfigure}%
\caption{Trajectories of the RC car obtained under different dynamical settings. Starting at rest, we apply the same control sequence for 50 timesteps at 30 Hz. We repeat the experiment three times for each setting and plot the mean trajectory. The crosses along the trajectory correspond to the car's mean position at an interval of 10 timesteps. The ellipses correspond to the empirical standard deviation in the car's position. The first two digits in the legend labels denote the sets of wheels used in the front and rear, respectively, and the third digit denotes the added weight in hectograms.}
\vspace{-15pt}
\label{fig:training_environments}
\end{figure}

\subsection{Hardware Experiments}
\begin{figure*}[t]
    \vspace{8pt}
    \centering
    \includegraphics[width=0.95\textwidth]{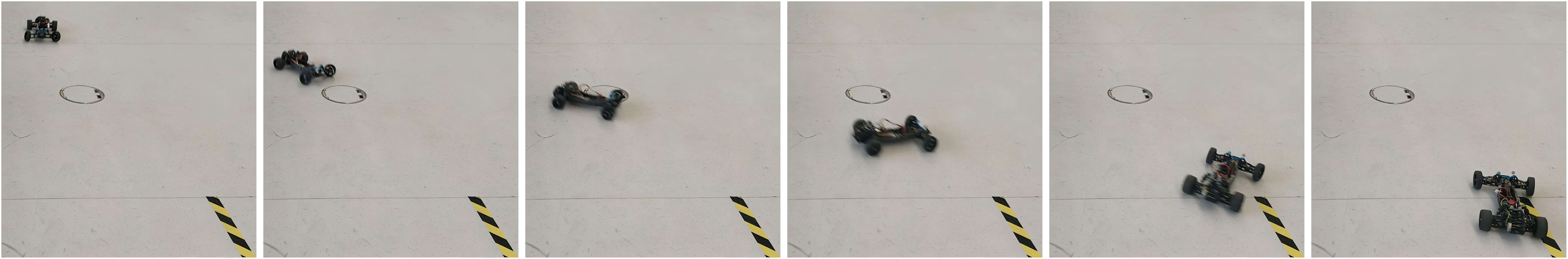}
    \caption{The RC Car in the process of performing a highly dynamic parking maneuver.}
    \label{fig:real_car_motion}
    \vspace{-12pt}
\end{figure*}

\looseness -1 We use a high-torque motor remote-controlled (RC) car \added{\citep[see][]{sukhija2023gradient} for hardware experiments. The car} can perform highly dynamic and nonlinear maneuvers such as drifting. The RL problem is to park reverse on a target position \added{that} is ca. 2 m away from the start position. This typically involves quickly rotating the car by 180$^{\circ}$ and then parking in reverse (see Fig.~\ref{fig:real_car_motion} or accompanying video\footnote{\href{https://crl.ethz.ch/videos/pacoh_model_based_rl.mp4}{https://crl.ethz.ch/videos/pacoh\_model\_based\_rl.mp4}}).
We represent the car with a six-dimensional state: the two-dimensional position and orientation of the car, and the corresponding velocities. The inputs to the car are the steering angle and throttle. 

To simulate different dynamical settings, we consider two different tire profiles with a varying amount of grip for both the front and rear wheels. Furthermore, we change the weight of the car from ca. \SI{1.6}{\kilo\gram} to \SI{1.8}{\kilo\gram}, and \SI{2}{\kilo\gram} by adding weights and operating the car in slow and fast mode. In the slow mode, the motor consumes less power and applies smaller accelerations. In contrast, in the fast mode, the car accelerates considerably faster and performs highly dynamic maneuvers such as drifting. The hardware setup is illustrated in Fig.~\ref{fig:rc_car_setup}.
In total, this gives us twenty-four settings, which result in considerable differences in the dynamic behavior of the car (see Fig.~\ref{fig:training_environments}). The dynamicity and variability of the different settings make the RC car a compelling platform for applying \algname.

\looseness -1 To collect meta-training datasets, we record trajectories under some of the different settings discussed above. In particular, for meta-training, we take only 5 tasks and use 4 minutes of recorded trajectories per task. After the meta-learning phase of PACOH-RL, we proceed with model-based RL on a new, highly dynamical target task. Instead of iCEM, we use SAC since the inference time of a SAC policy is much smaller than MPC, and, thus, can be run in real-time. This means that Line \ref{alg_line:mpc_rollout} in Alg.~\ref{alg:pacoh_rl} is replaced by training a NN policy with SAC~\cite{sac} on the current BNN dynamics model in a similar fashion to \cite{janner2021trust} and, then, running the trained policy on the real car to collect one trajectory.

\looseness - 1\textbf{Does \algname demonstrate sample efficiency on real-world hardware systems? }
To demonstrate the benefits of meta-learning, we compare \algname (Greedy) with PETS-DL~\cite{chua2018deep}. We use the greedy versions since the reward signal is dense, and incentivizing additional exploration does not help.
We chose one of the most dynamic settings of the RC car as the evaluation task, where we operate the car in fast mode with an added weight of \SI{0.2}{\kilo\gram} and the set of tires with the lowest friction. 
During the RL phase of both methods, we use SAC~\cite{sac} to train policies which are then deployed on the car. Overall, we ran the experiment for 20 episodes, each consisting of updating the BNN dynamics model with the latest data, SAC training of the policy, and collecting one trajectory on the car.

\begin{figure}[t]
    \centering
    \includegraphics[width=0.32\textwidth]{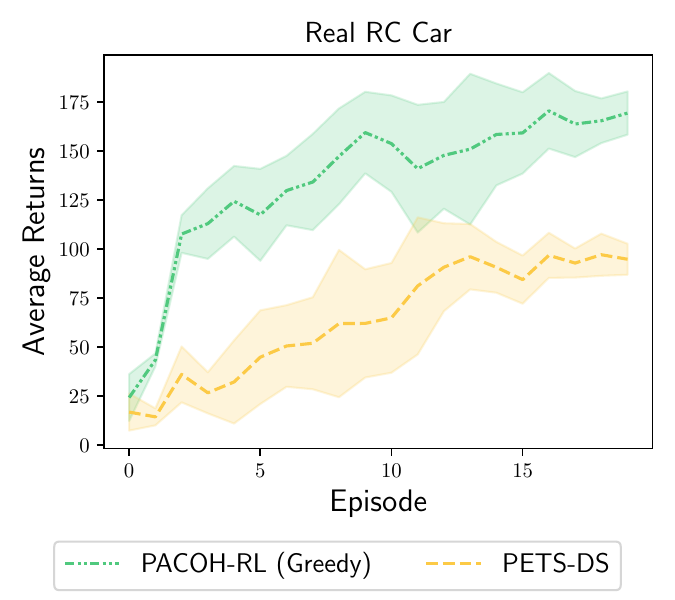}
    \caption{Returns on the real car averaged over three different seeds. We compare \algname to its non-meta-learning counterpart. \algname systematically outperforms the baseline in terms of sample efficiency and average return. \vspace{-10pt}}
    \label{fig:real_normal_test}
\end{figure}

Fig. \ref{fig:real_normal_test} displays the returns across the 20 episodes, averaged over 3 seeds.
We observe that \algname significantly outperforms the non-meta-learning approach. The \algname agent achieves high returns within the first few rollouts and converges to an almost optimal policy within less than 2 min (20 episodes) of real-world data. Due to its principled meta-level regularization, \algname successfully meta-learns a prior specific to RC car dynamics from only 5 meta-learning tasks. Equipped with this prior, the BNN dynamic model adapts quickly to the target task, resulting in substantial efficiency improvements compared to PETS.

\section{Conclusion}
\label{sec:conclusion}
\looseness -1 We have proposed \algname, a novel approach to model-based Meta-RL. \algname meta-learns a dynamics model prior from previous experience on the robotic platform, and, thereby, facilitates efficient adaptation under new dynamical conditions. The key characteristics of our method are its principled regularization and its holistic treatment of epistemic uncertainty, which guides exploration and data collection during the task adaptation stage. This allows \algname to achieve positive transfer and efficient task adaptation, even with only a handful of meta-learning datasets. Hence, \algname is one of the first Meta-RL approaches that are applicable to real robotic hardware where data is scarce.

\looseness -1 We have focused on harnessing the epistemic uncertainty quantification of our approach towards exploration. However, there remain many other relevant problems to explore that may benefit from meta-learned dynamics priors with reliable uncertainty estimates; for example, RL under safety constraints \citep{garcia2015comprehensive, as2022constrained} and off-policy evaluation \citep{rothfuss2023hallucinated, hanna2017bootstrap}.

\bibliography{refercences}
\ifthenelse{\boolean{ArXivSubmission}}{

\appendices
\section{Method Details}
\label{sec:method_details}

\subsection{Meta-Learning Dynamics Model Priors}

In the meta-learning stage, PACOH-RL employs the \textsc{PACOH-NN} approach of \cite{rothfuss2021pacoh, rothfuss2022pac}. We are given $n$ datasets of transition data $\calD_1, ..., \calD_n$ where each dataset $\calD_i = \{ (s, a, s') \}$ contains $m_i = |\calD_i|$ transition triplets corresponding to MDP $\calT_i$. In addition, the meta-learner presumes a hyper-prior over the prior parameters $\phi$ which we choose to be a zero-centered Gaussian $\mathcal{P} = \calN(0, \sigma_\calP^2)$ with variance vector $\sigma_\calP^2 \in \R^{\text{dim}(\phi)} = \R^{2 \text{dim}(\theta)}$. 

We initialize the $K=3$ prior particles $\phi_{1}, ... \phi_K$ by sampling i.i.d. from the hyper prior, i.e. $\phi_k \sim \calP$. Then, we perform SVGD \cite{liu2016stein} to approximate the hyper-posterior 
$\mathcal{Q}(\phi)$ in (\ref{eq:pacoh}).  As kernel function for SVGD, we use a squared exponential kernel with length scale (hyper-)parameter $\ell$, i.e., $k(\phi ,\phi') = \exp \left( - \frac{|| \phi - \phi'||_2^2}{2 \ell} \right)$. In each iteration, we first sample a batch of $n_b \leq n$ meta-training datasets, and from each dataset in the batch, we sample a batch of $m_b \leq m_i$ data points. The denote the resulting batches as $\tilde{\calD}_1, ..., \tilde{\calD}_{n_b}$. Note that the indexing here is distinct from before. Then, for each prior particle $P_{\phi_k}$ we draw $L=5$ NN parameters, i.e., $\theta_1, ..., \theta_L \sim P_{\phi_k}$ to estimate the (generalized) marginal log-likelihood (MLL) via
\begin{equation} \label{eq:mll_estimator}
\ln \tilde{Z}(\calD_i, P_\phi) := ~ \text{LSE}_{l=1}^L\left( - \sqrt{m_i}  \hat{\calL}(\theta_l, \tilde{\calD}_i) \right) - \ln L 
\end{equation} 
where $\text{LSE}$ is the Log-Sum-Exp function and $\hat{\calL}(\theta_l, \tilde{\calD}_i) = \frac{1}{|\tilde{\calD}_i|} \sum_{(s, a, s') \in \tilde{\calD}_i} \ln p(s'|s, a, \theta_l)$ is the average log-likelihood for the transition triplets in the batch $\tilde{\calD}_i$, corresponding to the NN with parameters $\theta_l$. With these MLL estimates, we can compute an estimate of the hyper-posterior score:
\begin{equation}
    \nabla_{\phi} \ln \tilde{\calQ}^*(\phi) := \nabla_{\phi} \ln \calP(\phi) + \frac{n}{n_{bs}}\sum_{i=1}^{n_{bs}} \frac{1}{\sqrt{n m_i} + 1} \nabla_{\phi} \ln \tilde{Z}(S_i, P_{\phi})
\end{equation}
which we then use to update the SVGD prior particles via the SVGD update rule, i.e., $ \forall{k \in [K]} $,
\begin{equation}
    \phi_k \leftarrow \phi_k + \frac{\eta}{K} \sum_{k'=1}^K \left [k(\phi_{k'}, \phi_k) \nabla_{\phi_{k'}} \ln \tilde{\calQ}^*(\phi_{k'}) +  \nabla_{\phi_{k'}} k(\phi_{k'}, \phi_k) \right] ~.
\end{equation}
The PACOH-NN procedure is summarized in Algorithm \ref{alg:pacoh_nn}. After convergence, it returns the set of priors  $\{ P_{\phi_1}, ..., P_{\phi_K} \}$ which approximate the PAC-Bayesian hyper-posterior $\calQ(\phi)$.

\begin{algorithm*}[t]
\caption{\textsc{PACOH-NN}}\label{alg:pacoh_nn}
\hspace*{\algorithmicindent} \textbf{Input:}  Datasets $\calD_1, ..., \calD_n$, hyper-prior $\calP$, step size $\eta$, number of particles $K$
\begin{algorithmic}[1]
\While{not converged} 
\State $\{ \phi_1, ..., \phi_K\} \sim \calP$ \Comment{Sample $K$ prior particles from hyper-prior}
    \State $\{\calD_1, ..., \calD_{n_b}\} \leftarrow$ Sample batch of $n_{b}$ tasks from meta-training datasets 
 	\For{$i=1, ..., n_{b}$}
 	    \State $\tilde{\calD}_i \leftarrow$ Sample batch of $m_{b}$ data points from $\calD_i$
 	\EndFor
 	\For{$k=1,...,K$} 
     	\State $\{\theta_1, ..., \theta_L\} \sim P_{\phi_k}$ \Comment{sample NN-parameters from priors}
     	\For{$i=1, ..., n_{bs}$}
     	    \State $\ln \tilde{Z}(\tilde{\calD}_i, P_{\phi_k}) \leftarrow \text{LSE}_{l=1}^L\left( - \beta_i  \hat{\calL}(\theta_l, \tilde{\calD}_i) \right) - \ln L$ \hfill \Comment{estimate generalized MLL}
     	\EndFor
     	\State $  \nabla_{\phi_k} \ln \tilde{\calQ}^*(\phi_k)  \leftarrow \nabla_{\phi_k} \ln \calP(\phi_k) + \frac{n}{n_{bs}}\sum_{i=1}^{n_{bs}} \frac{\1}{\sqrt{nm_i} + 1} \nabla_{\phi_k} \ln \tilde{Z}(S_i, P_{\phi_k})$ \hfill \Comment{score}
	\EndFor
	\State $\phi_k \leftarrow \phi_k + \frac{\eta}{K} \sum_{k'=1}^K \left [k(\phi_{k'}, \phi_k) \nabla_{\phi_{k'}} \ln \tilde{\calQ}^*(\phi_{k'}) +  \nabla_{\phi_{k'}} k(\phi_{k'}, \phi_k) \right] \forall{k \in [K]}$  \Comment{SVGD}
\EndWhile
\hspace*{\algorithmicindent} \textbf{Output:}  set of priors $\{ P_{\phi_1}, ..., P_{\phi_K} \}$ as approximation of the hyper-posterior $\mathcal{Q}$
\end{algorithmic}
\end{algorithm*}

\hspace{5pt}
\subsubsection{Adapting the dynamics model to the target task}

\begin{algorithm*}
\caption{BNN-SVGD}
\label{algo:svgd_bnn}
\hspace*{\algorithmicindent} \textbf{Inputs:} BNN prior $P_{\phi_k}$, target training dataset $\calD^*$ \\
\hspace*{\algorithmicindent} \textbf{Parameters:} Kernel function $k(\cdot, \cdot)$, SVGD step size $\nu$, number of particles $L$ 
\begin{algorithmic}[1]
 	\State $\{\theta^k_1, ..., \theta^k_L\} \sim P_{\phi_k}$ \hfill \Comment{initialize NN posterior particles from $k$-th prior}
\While{not converged} 
    \For{$l=1,...,L$}
    \State $\nabla_{\theta^k_l} \ln Q^*(\theta^k_l))  \leftarrow \nabla_{\theta^k_l} \ln P_{\phi_k}(\theta^k_l)) + \beta ~ \nabla_{\theta^k_l}  \calL(l, \tilde{S})$ \Comment{compute posterior score}
    \EndFor
    \State $\theta^k_l \leftarrow \theta^k_l +  \frac{\nu}{L} \sum_{l'=1}^L \left [k(\theta^k_{l'}, \theta^k_l) \nabla_{\theta^k_{l'}} \ln Q_k(\theta^k_{l'}) +  \nabla_{\theta^k_{l'}} k(\theta^k_{l'}, \theta^k_l) \right] \forall{l \in [L]}$ 
\EndWhile
\end{algorithmic}
\hspace*{\algorithmicindent} \textbf{Output:} Set of NN parameters $\Theta_k = \{\theta_1^k\, ..., \theta_L^k\}$
\end{algorithm*}

After the meta-learning stage, PACOH-RL performs model-based RL on the target task, using the meta-learned BNN prior for the dynamics model. In every episode, we update the BNN dynamics model to also incorporate the latest trajectory as training data (c.f. line \ref{alg_line:svgd1}  and \ref{alg_line:svgd2} of Algorithm \ref{alg:pacoh_rl}). We combine the latest training datasets $\calD^*$ with the meta-learned priors $P_{\phi_k}, k=1, ..., n$, via the the (generalized) Bayesian posterior
\begin{equation}
    Q_k(\theta; \calD^*) \propto p(\calD^*|\theta)^{1/\sqrt{n}} P_{\phi_k}(\theta) ~.
\end{equation} 
We approximate each posterior via SVGD \cite{liu2016stein} with an SE kernel function $k(\cdot, \cdot)$.

In particular, for each prior, we first initialize $L=5$ NN particles $\{\theta^k_1, ..., \theta^k_L\} \sim P_{\phi_k}$ by sampling from the the prior $P_{\phi_k}$. Then, we compute the posterior score 
\begin{equation}
    \nabla_{\theta^k_l} Q_k(\theta^k_l))  \leftarrow \nabla_{\theta^k_l} \ln P_{\phi_k}(\theta^k_l)) + \sqrt{m^*} ~ \nabla_{\theta^k_l}  \hat{\calL}(\theta^k_l, \calD^*)
\end{equation}
where $m^* = |\calD^*|$ is dataset size and 
$\hat{\calL}(\theta^k_l, \calD^*) = \frac{1}{m^*} \sum_{(s, a, s') \in \calD^*} \ln p(s'|s, a, \theta_l)$ is the average log-likelihood of the NN with parameters with $\theta_l$ on $\calD^*$. Based on the posterior score, we can update the NN particles via SVGD as follows:
\begin{equation}
    \theta^k_l \leftarrow \theta^k_l +  \frac{\nu}{L} \sum_{l'=1}^L \left [k(\theta^k_{l'}, \theta^k_l) \nabla_{\theta^k_{l'}} \ln Q_k(\theta^k_{l'}) +  \nabla_{\theta^k_{l'}} k(\theta^k_{l'}, \theta^k_l) \right] ~.
\end{equation}

We repeat these steps until the set of NN particles converges.
This BNN-SVGD approximate inference procedure is summarized in \ref{algo:svgd_bnn}. The result of each the BNN-SVGD runs for a prior $P_{\phi_k}$ is a set of NN parameters $\Theta_k = \{\theta_1^k\, ..., \theta_L^k\}$ which approximate the posterior $Q_k$. We aggregate the NN parameters via $\Theta := \{\Theta_1, ..., \Theta_K \}$ which corresponds to $K \cdot L = 15$ neural networks.

When making dynamics predictions, we aggregate the different neural networks predictions into a Gaussian approximation $$\hat{p}_{\Theta}(s'| s, a) = \calN(s' ; \hat{\mu}_{\Theta}(s, a), \hat{\sigma}^2_{\Theta}(s, a))$$ where $$\hat{\mu}_{\Theta}(s, a) = \frac{1}{KL} \sum_{k=1}^K \sum_{l=1}^L  h_{\theta_{k,l}}(s, a)$$ is the predictive mean and $$\hat{\sigma}^2_{\Theta}(s, a) = \frac{1}{KL} \sum_{k=1}^K \sum_{l=1}^L \left(h_{\theta_{k,l}}(s, a) - \hat{\mu}(s, a)\right)^2$$ the epistemic variance.

\subsection{Model-based Control of PACOH-RL}

In Section \ref{sec:mb_control}, we proposed using either MPC or SAC to solve the optimistic control problem in  \Cref{eq:hucrl_policy_optimization}. In this section, we provide additional details about these two variants of PACOH-RL.

\vspace{5pt}
\subsubsection{The Model Predictive Control Variant}  \mbox{}

\textbf{Background on MPC:}
Model predictive control (MPC) \cite{camacho2013model} is a control strategy that uses a dynamics model to predict the future behavior of a system/MDP. At every step, the MPC controller plans an optimal action sequence for a horizon of $H$ steps. If the dynamics model is a deterministic function $\hat{f}(s, a) \mapsto s'$, the optimal $H$-step action sequence is the solution to the following optimal control problem:
\begin{equation} \label{eq:mpc_det}
    a^*_t, ..., a^*_{t+H} = \argmax_{a_t, ..., a_{t+H}} \sum_{t'=t}^{t+H} r(s_{t'}, a_{t'})  ~~~ \text{where} ~~ s_{t'+1} = \hat{f}(s_{t'}, a_{t'}) ~.
\end{equation}
If the dynamics model of probabilistic, i.e., a condition distribution $\hat{p}(s'|s, a)$, it induces a probability distribution over sequences. Hence, the MPC controller aims so solve the planning problem in expectation, i.e.,
\begin{equation} \label{eq:mpc_prob}
    a^*_t, ..., a^*_{t+H} = \argmax_{a_t, ..., a_{t+H}} \E \left[ \sum_{t'=t}^{t+H} r(s_{t'}, a_{t'}) \right] ~,~~ s_{t'+1} \sim \hat{p}(s'|s, a) ~.
\end{equation}
Once the controller has obtained the optimal action sequence, it executes the first action $a^*_t$ in the sequence. After taking into consideration the observation of the new state, the process is repeated.

\textbf{Cross-Entropy Method.:} The cross-entropy method (CEM) \cite{cem} is a black-box optimization approach that can be used to solve the MPC planning problems in (\ref{eq:mpc_det}) and (\ref{eq:mpc_prob}).
It optimizes the action sequence by sampling and evaluating (calculating the reward) multiple candidate sequences. It uses a probabilistic approach to gradually refine the distribution of the action sequences, favoring those with higher rewards. Typically, a set of elite sequences with the highest reward is used to fit a Gaussian distribution, which is used in the next iteration to sample candidates. This leads to incremental improvements in the action sequence candidates. After a pre-defined number of iterations of sampling candidates and re-fitting the sampling distribution on the elites, the algorithm returns the best action sequence from the current candidate sequences.

\begin{algorithm*}
\caption{\textsc{iCEM-MPC}: Improved Cross-Entropy Method for Optimistic MPC}\label{alg:icem}
\hspace*{\algorithmicindent} \textbf{Input:} Dynamics model $\hat{p}_\Theta$, MDP $\calT = (\mathcal{S}, \mathcal{A}, p, p_0, r, T)$ \\
\hspace*{\algorithmicindent} \textbf{Parameters:} Number of CEM iterations  $n_{it}$, number of particles $n_p$, planning horizon $h$, \\
\hspace*{\algorithmicindent} \textbf{Parameters:} Number of elites $n_e$, reduction factor $\gamma$, initial sample variance $\sigma_{\textnormal{init}}$

\begin{algorithmic}[1]
\For {$t=0, ..., T-1$}
  \If {$t = 0$}
    \State $\mu_0 \gets \mathbf{0} \in \R^{(d_a + d_s) \times h}$  \Comment{Initialize mean} 
  \Else
    \State $\mu_t \gets \text{shifted} ~ \mu_t ~ \text{(with last time-step repeated)}$
  \EndIf
  \State $\sigma_t \gets \sigma_{\textnormal{init}} \cdot \mathbf{1} \in \R^{(d_a + d_s) \times h}$ \Comment{Initialize standard deviation} 
  \For{$i=1,2,\ldots,n_{it}$}
    \State $n_{p,i} \leftarrow \max(n_p\cdot\gamma^{-i},2\cdot n_e)$ \Comment{Update number of particles}
    \State $ \text{samples} \leftarrow n_{p,i}$ samples from $\textnormal{clip}(\mu_t + C^\beta(d_a+d_s, h) \odot \sigma_t^2)$
    \If{$i = 0$} 
      \State add fraction of shifted $\text{elite-set}_t$ to $\text{samples}$
    \Else
      \State add fraction of $\text{elite-set}_t$ to $\text{samples}$
    \EndIf
    \If{$i = n_{it}$}
      \State add $\mu_t$ to \text{samples} 
    \EndIf
      \For{$A \in \text{samples}$}
        \State $\tau, R \gets$ \textsc{SimulateOptimisticTrajectory}($\hat{p}_\Theta$, $s^*_t$, $r$, $H$, $A$) \Comment{Algorithm \ref{alg:trajectory_sim}} \label{alg_line:opt_traj_sim}
      \EndFor
  \State $\text{elite-set}_t \gets$ $n_e$ action sequences with highest return $R$
  \State $\mu_t, \sigma_t \gets $ fit Gaussian distribution to $\text{elite-set}_t$
  \EndFor
  \State execute first action $a_t^*$ of the best elite sequence on MDP $\calT$ and observe next state $s^*_{t+1}$
\EndFor
\end{algorithmic}
\hspace*{\algorithmicindent} \textbf{Returns:} Executed trajectory $(s^*_0, a^*_0, ...., s^*_{T-1}, a^*_{T-1}, s^*_T)$
\end{algorithm*}

\textbf{Improved Cross-Entropy Method:} Improved cross-entropy method (iCEM) was proposed by \cite{icem} as a refined version of the cross-entropy method with a focus on efficiently solving control problems. The improved cross-entropy method uses the following modifications w.r.t. to the CEM method
\begin{itemize}
    \item It uses the fitted distribution from the previous timestep to sample initial candidate sequences.
    \item It uses colored noise to sample from the fitted distribution. This results in auto-correlated action sequences that allow for more directed exploration behavior.
    \item It adds a fraction of the elite sequences to the candidate action sequences in the next iteration.
\end{itemize}

Algorithm \ref{alg:icem} summarizes how we use iCEM for optimistic MPC. There, $C^\beta(d, h)$ denotes the colored noise sampling function that returns d (one for each action dimension) sequences of length h (horizon) sampled from colored noise distribution with exponent $\beta$ and with zero mean and unit variance. For details about sampling colored noise sequences, we refer to \cite[][Section 3.1]{icem}.

\textbf{Uncertainty-Aware Optimistic Control with MPC:} We aim to solve the uncertainty-aware, optimistic control problem in \Cref{eq:hucrl_policy_optimization} with MPC. Hence, at every timestep $t$, we plan an augmented action sequence $(a_t, \eta_t, ..., a_{t+H}, \eta_{t+H})$ where $\eta \in [-1, 1]^{\text{dim}(\calS)}$ are the hallucinated controls that optimistically choose one plausible state transition from the (epistemic) confidence region $[\hat{\mu}(s_t, a_t) \pm  \nu \hat{\sigma}(s_t, a_t)]$ of our dynamics model. The corresponding MPC optimization problem follows as:
\begin{equation} \label{eq:mpc_hucrl_objective}
\begin{split}
    &\argmax_{a_t, \eta_t, ..., a_{t+H}, \eta_{t+H}} ~ \sum_{t'=t}^{t+H} r(s_{t'}, a_{t'})   ~ \\ ~ \text{ s.t. }  &s_{t'+1} = \hat{\mu}(s_{t'}, a_{t'}) + \nu \eta_{t'} \hat{\sigma}(s_{t'}, a_{t'}) ~
\end{split}
\end{equation}
We solve (\ref{eq:mpc_hucrl_objective}) with iCEM. The overall procedure of rolling out a trajectory with iCEM-MPC is summarized in Algorithm \ref{alg:icem}. In line \ref{alg_line:opt_traj_sim}, we simulate optimistic trajectories with the dynamics model and the candidate augmented control sequences. This is further detailed in Algorithm \ref{alg:trajectory_sim}.

\begin{algorithm*}
\caption{\textsc{SimulateOptimisticTrajectory}}\label{alg:trajectory_sim}
\hspace*{\algorithmicindent} \textbf{Input:} Dynamics model $\hat{p}_\Theta$, Current state $s_t$, reward function $r$ \\
\hspace*{\algorithmicindent} \textbf{Input:}  Horizon $H$, augmented action sequence $A=(a_t, \eta_t, ..., a_{t+H}, \eta_{t+H})$
\begin{algorithmic}[1]
\State $R \leftarrow 0$ \Comment{Initialize return to zero}
\For {$h=0, ..., H$}
  \State $s_{t+h+1} \leftarrow \hat{\mu}_{\Theta}(s_{t+h}, a_{t+h}) + \nu \eta_{t+h} \hat{\sigma}_{\Theta}(s_{t+h}, a_{t+h})$ \Comment{Compute next state}
  \State $R \leftarrow R + r(s_{t+h}, a_{t+h})$ \Comment{Add reward to return}
\EndFor
\end{algorithmic}
\hspace*{\algorithmicindent} \textbf{Returns:} Simulated trajectory $\tau=(s_t, a_t, ..., s_{t+H}, a_{t+H})$, Return $R$ of trajectory 
\end{algorithm*}

\vspace{5pt}
\subsubsection{The Soft-Actor Critic Variant} \mbox{}
\label{ssec:sac}

\textbf{Background on Soft Actor-Critic:}
Soft Actor-Critic (SAC) \cite{sac} is a widely used off-policy algorithm and has been empirically shown to work well on a wide variety of continuous-control RL problems. It uses a maximum entropy reinforcement learning setting, where the agent aims to maximize simultaneously the rewards and the entropy of the learned policy. It builds upon the formulation of a soft-MDP, where the RL objective is a combination of the returns and the conditional entropy $ H(\pi_\vartheta(a|s))$ of the policy:
\begin{equation}
J(\vartheta) = \sum_{t=0}^T \mathbb{E}_{s_t \sim \rho_{t, \pi_\vartheta}} \mathbb{E}_{a_t \sim \pi_\vartheta(a_t, s_t)} \left[ r(s_t, a_t) + \lambda H(\pi_\vartheta(a_t|s_t)) \right],
\end{equation}
Here $\pi_\vartheta$ is a parameterized (neural network) policy with parameters $\vartheta$ and $\rho_{t, \pi_\vartheta}$ is the state-occupancy measure at step $t$. To optimize $J(\vartheta)$, SAC uses (soft) critics, in particular, a value- and a Q-function. In practice, to train SAC, we follow a similar approach to \cite{janner2021trust}. 

\textbf{Uncertainty-Aware Optimistic Control with SAC:}
We aim to solve the uncertainty-aware, optimistic control problem in \Cref{eq:hucrl_policy_optimization} with SAC. For that, we use a neural network policy for both the (real) actions $a \in \R^d_a$ and the hallucinated controls $\eta \in \R^{d_s}$, i.e., $\pi_\vartheta(a, \eta | s)$. We parametrize the policy as conditional Gaussian $\calN(\mu_\vartheta(s), \sigma^2_\vartheta(s))$ where $\mu_\vartheta(s), \sigma^2_\vartheta(s) \in \R^{d_s + d_a}$ are outputs of the neural network which takes the current state $s$ as an input.

In every episode, we train the policy $\pi_\vartheta$ with SAC on the hallucinated, optimistic transition model
\begin{equation}
    s' = f(s, \eta, a) = \hat{\mu}_{\Theta}(s, a) + \nu \eta \hat{\sigma}_{\Theta}(s, a) ~.
\end{equation}
That is, SAC uses $f(s, \eta, a)$ to generate rollouts/state transitions to fill the replay buffer. When performing rollouts with the policy $\pi_\vartheta$ on the real environment/MDP, the hallucinated controls $\eta$ are simply ignored.

Algorithm \ref{alg:pacoh_rl_sac} summarizes the PACOH-RL version with SAC policy. Initially, we initialize the policy $\pi_\vartheta$ randomly (line \ref{alg-line:policy_init}). Then, in every episode, we train the policy on the optimistic dynamics model based on the updated BNN (line \ref{alg-line:policy-training}), and, then, roll out one trajectory with the trained policy $\pi_\vartheta$ on the real environment, i.e., the target task $\mathcal{T}^*$ (line \ref{alg_line:policy_rollout}).

\begin{algorithm*}
\caption{PACOH-RL (SAC version)}\label{alg:pacoh_rl_sac}
\hspace*{\algorithmicindent} \textbf{Input:} 
Tansitions datasets $\{\calD_1,\ldots, \calD_n\}$ from previous tasks, test task $\calT^{*}$, hyper-prior $\calP$

\begin{algorithmic}[1]
\State $\{P_{\phi_1}, \ldots, P_{\phi_K} \} \leftarrow \textsc{PACOH-NN}(\calD_1,\ldots, \calD_n, \calP)$ \hfill \Comment{Meta-learn set of priors to approx. $\calQ$}
\State $\calD^* \leftarrow \emptyset$  \Comment{Initialize empty transition dataset}
\State $\pi_\vartheta \gets$ Initialize policy \label{alg-line:policy_init}
\For{$\textnormal{episode}=1,2,\ldots$}
    \For{$k=1,2,\ldots,K$}
        \State $\Theta_k \leftarrow \textsc{SVGD}(\calD^*, P_{\phi_k})$  \Comment{Train BNN with latest transition data}
    \EndFor
    \State $\Theta := \{ \Theta_1, ..., \Theta_K\}$
    \State $\hat{p}_\Theta \gets \calN(\hat{\mu}_{\Theta}(s, a), \hat{\sigma}^2_{\Theta}(s, a))$ \Comment{Aggregate NN predictions into predictive distribution}
    \State $\pi_\vartheta \gets \textsc{SAC}(\pi_\vartheta, \hat{p}_\Theta)$ \Comment{Train policy on optimistic dynamics model} \label{alg-line:policy-training}
\State $(s_0, a_0, ..., a_{T-1}, s_T) \leftarrow$ Execute trajectory with $\pi_\vartheta$ on $\calT^*$ \Comment{Run policy on target task} \label{alg_line:policy_rollout}
\State $\calD^* \leftarrow \calD^* \cup \{(s_t, a_t, s_{t+1})\}_{t=0}^{T-1}$  \Comment{Add transitions to dataset}
\EndFor
\end{algorithmic}
\end{algorithm*}

\vspace{5pt}
\subsubsection{The Greedy Variant} \mbox{}
\label{ssec:greedy}

In \Cref{sec:experiments}, we introduce PACOH-RL (Greedy), a greedy version of PACOH-RL. It is based on the distributional sampling method proposed by~\cite{chua2018deep}.  In the planning, we sample the next state from a Gaussian distribution induced by the predictive mean and epistemic variance, i.e., $\calN(\hat{\mu}_{\Theta}(s_t, a_t), \hat{\sigma}_{\Theta}^2(s, a))$. This results in the following policy optimization;
\begin{equation} \label{eq:neutral_policy_optimization}
\begin{split}
    \pi&^{\text{DS}}  = \argmax_\pi ~ \E_{a_t \sim \pi(a_t | s_t)} \left[ \sum_{t=0}^{T-1} r(s_t, a_t) \right] 
    \\ &\text{ s.t. } s_{t+1} \sim \calN(\hat{\mu}_{\Theta}(s_t, a_t), \hat{\sigma}_{\Theta}^2(s_t, a_t)).
\end{split}
\end{equation}
The proposed planning method is robust w.r.t. the epistemic uncertainty in the transition model; however, it does not leverage the epistemic uncertainty for guiding exploration compared to the policy optimization in \Cref{eq:hucrl_policy_optimization}. Accordingly, in environments with scarce rewards, PACOH-RL performs considerably better than its greed variants (see \Cref{sec:experiments}).

\section{Experimental Details} \label{sec:details_exps}

\subsection{Hardware Experiments}

We control the car at \SI{30}{\hertz} and use the Optitrack for robotics motion capture system
\footnote{\href{https://optitrack.com/applications/robotics/}{https://optitrack.com/applications/robotics/}}
to obtain position estimates of the car at \SI{120}{\hertz}. We estimate the velocities with finite differences and apply a moving average filter with a window size of six-time steps.
Since the transmission and execution of the control signals on the car is delayed by ca. 70 - 80 ms, the current change $s_t \rightarrow s_{t+1}$ in the car's state is mainly governed by $a_{t-3}$. Hence, we append the last three actions $[a_{t-3}, a_{t-2}, a_{t-1}]$ to the current state $s_t$.
We use the same reward function for both

\subsection{Hyperparameters.}
We list all hyperparameters used for training the BNN models and PACOH meta-learner in \Cref{tab:model_params}. The parameters for iCEM are listed in \Cref{tab:controller_settings}.

\paragraph{Meta-Learning Data.}
As previously discussed, PACOH-RL is a two-phase procedure. We first collect data $\{\calB_1,\ldots, \calB_n\}$ that corresponds to a sequence of tasks $\calM_1, ..., \calM_n \sim p(\calM)$ and used the PACOH-NN algorithm to approximate hyper-posterior by a set of prior particles $\{P_{\phi_1}, \ldots, P_{\phi_K} \}$. We then use these meta-learned prior particles for the BNN models representing the dynamics of $\calM^{*}$. In practice, to collect each meta-learning dataset $\calB_i$, we roll out the policy for $E$ episodes for each task. In \Cref{tab:data_collection}, we summarize the number of tasks and episodes used during the meta-training phase of PACOH-RL.

\paragraph{Reward Functions.}
For the non-sparse reward environments Pendulum, Cartpole, and Half-Cheetah from OpenAI gym, we use the same reward functions used in the standard Gym API \cite{OpenAIGym}. For the simulated RC Car and the real-world experiments, we use a modification of the `tolerance' reward, introduced in the DeepMind Control Suite~\cite{tassa2018deepmind}. Sparse rewards for the environments Pendulum, Cartpole, and Pusher are also implemented using the `tolerance' function. The exact forms of the reward functions are listed in \Cref{tab:reward_funcs}. All environments have an associated action cost, which is just the action cost factor times the square of the $l^{2}$-norm of the action.

\begin{table}[h]
    \centering
    \caption{Model Hyperparameters}
    \label{tab:model_params}
    \begin{subtable}[t]{.46\textwidth}
        \centering
        \caption{Hyperparameters for BNN model.}
        \begin{adjustbox}{max width=\linewidth}\begin{threeparttable}
        \begin{tabular}{@{}ll@{}}
        \toprule
         \textbf{Parameter} & \textbf{Value} \\
 \hline
 Layers & $[200] \times 4$\\
 Non-linearity & ReLU \\
 Learning rate & 0.001 \\
 Training steps & 2000 \\
 Batch size & 32 \\
 Likelihood standard deviation ($\sigma_y$) & 0.1\\
 Kernel function ($k(\cdot, \cdot)$) & RBF\\
 Kernel function bandwidth & 10.0\\
 Default prior weight distribution & $\mathcal{N}(\mathbf{0}, 0.1\cdot\mathbb{I})$ \\
 Default prior likelihood distribution & $\mathcal{N}(\ln(0.1)\cdot\mathbf{1}, \mathbb{I})$ \\
        \bottomrule
        \end{tabular}
        \end{threeparttable}\end{adjustbox}
\end{subtable}
\newline
\vspace*{9pt}
\newline
\begin{subtable}[t]{.5\textwidth}
        \centering
        \caption{Hyperparameters for PACOH meta-learner.}
          \begin{adjustbox}{max width=\linewidth}\begin{threeparttable}
        \begin{tabular}{@{}ll@{}}
        \toprule
         \textbf{Parameter} & \textbf{Value} \\
 \hline
 Meta training steps ($l$) & 100000\\
 Learning Rate ($\eta$) & 0.0008 \\
 Number of prior particles ($K$) & 3 \\
 Number of model samples ($L$) & 3 \\
 Batch size ($n_b$) & 4 \\
 Samples per task ($m_b$) & 8 \\
 Kernel function ($k(\cdot, \cdot)$) & RBF \\
 Kernel function bandwidth & 10.0\\
 Weights mean hyper-prior & $\mathcal{N}(\mu_w; \textbf{0}, 0.4\cdot\mathbb{I})$ \\
 Weights standard deviation hyper-prior & $\mathcal{N}(\sigma_w; -3\cdot\mathbb{I}, 0.4\cdot\mathbb{I})$ \\
 Likelihood mean hyper-prior & $\mathcal{N}(\mu_{\sigma_y}; -8\cdot\mathbb{I}, \mathbb{I})$\\
 Likelihood standard deviation hyper-prior & $\mathcal{N}(\sigma_{\sigma_y}; -4\cdot\mathbb{I}, 0.2\cdot\mathbb{I})$\\
        \bottomrule
        \end{tabular}
\end{threeparttable}\end{adjustbox}
    \end{subtable}%
\end{table}

\begin{table}[]
    \centering
    \caption{iCEM parameters used for planning.}
    \label{tab:controller_settings}
    \begin{subtable}[t]{.48\textwidth}
        \centering
        \begin{tabular}{@{}ll@{}}
        \toprule
 \textbf{Parameter} & \textbf{Value} \\
 \hline
 Shooting method & iCEM\\
 Number of iterations ($n_{it}$) & 5 \\
 Number of particles ($n_p$) & 1000 \\
 Planning horizon ($H$) & 40 \\
 Number of elite samples ($n_e$) & 50 \\
 Reduction factor ($\gamma$) & 1.25 \\
 Initial sample variance ($\sigma_{\textnormal{init}}$) & 0.5 \\
 Colored noise beta ($C^{\beta}$) & 2.0 \\
 Distribution update factor ($\alpha$) & 0.2 \\
        \bottomrule
        \end{tabular}
    \end{subtable}%
\end{table}

\begin{table}[]
	\caption{Meta-Training Data collection}
	\label{tab:data_collection}
	\begin{center}
        \begin{tabular}{p{0.12\textwidth}>{\centering\arraybackslash}m{0.085\textwidth}>{\centering\arraybackslash}m{0.0879\textwidth}>{\centering\arraybackslash}m{0.07\textwidth}}
		\toprule
		\textbf{Domain}               & \textbf{Training tasks ($n$)}  & \textbf{Episodes per task ($E$)} &  \textbf{Total steps}                                        \\ \midrule
		\multicolumn{4}{c}{Simulation, dense rewards}                                                                                \\ \midrule
		Pendulum                & 20              & 40             &         200000                  
            \\
		Cartpole                & 15              & 10             &         30000 
            \\
            Half-Cheetah            & 20              & 60             &         1200000
            \\
            Simulated RC Car        & 20              & 25             &         100000
            \\ \midrule
            \multicolumn{4}{c}{Simulation, sparse rewards} 
            \\ \midrule
            Sparse Pendulum         & 20              & 40             &         200000 
            \\
            Sparse Cartpole         & 15              & 10             &         30000
            \\
            Sparse Pusher           & 20              & 25             &         120000
            \\ \midrule
		\multicolumn{4}{c}{RC Car}                                                                                                   \\ \midrule
            Parking                & 5              & 40             &            40000                                                           
            \\ \bottomrule
	\end{tabular}
	\end{center}
 \vspace{-9pt}
\end{table}

\begin{table*}[h]
\renewcommand{\arraystretch}{1.2}
  \caption{Reward functions used for different environments. $tol$ refers to the `tolerance' function from DM control suite~\cite{tassa2018deepmind}.}
	\label{tab:reward_funcs}
        \begin{tabular}{p{0.12\textwidth}>{\centering\arraybackslash}m{0.205\textwidth}>{\centering\arraybackslash}m{0.14\textwidth}p{0.44\textwidth}}
		\toprule
		\textbf{Domain}         & \textbf{Reward Function}   & \textbf{Action Cost Factor} &  \multicolumn{1}{c}{\textbf{Description}}\\ \midrule
		\multicolumn{4}{c}{Simulation, dense rewards}             \\ \midrule
		Pendulum                &          $ - \theta^{2} - 0.1 \omega ^ 2$   &    0.001        &  $\theta$: angle to goal position, $\omega$: angular velocity            
            \\
		Cartpole                &         $ - d ^ {2} / l ^ 2$  &    0.01    &  $d$: distance to goal position, $l$: length of the pole 
            \\
            Half-Cheetah            &         $ v_x $         &   0.1        & $v_x$: velocity along +x axis
            \\
            Simulated RC Car        &         $ tol(d) +
                                            0.5 * tol(\theta) $  & 0.05 &  $d$: distance to goal position, $\theta$: angular deviation from goal, \ \ \ \ \ \ \ \ \ \ \ \ \ \ \ \ $tol$ params - bounds: (0, 0.1), margin: 0.5, value at margin: 0.2
            \\ \midrule
            \multicolumn{4}{c}{Simulation, sparse rewards} 
            \\ \midrule
            Sparse Pendulum         &         $tol(\theta) + tol(\omega)$  &   0.001  &   $\theta \ tol$ params - bounds: (0.95, 1), margin: 0.3, value at margin: 0.1, $\omega \  tol$ params - bounds: (-0.5, 0.5), margin: 0.5, value at margin: 0.1
            \\
            Sparse Cartpole         &        $tol(\theta)$ & 0.01 & $\theta \  tol$ params - bounds: (0.995, 1), margin: 0.0, value at margin: 0.1
            \\
            Sparse Pusher           &         $tol(d_{e}) * (0.5 + 0.5 * tol(d_{g}))$ & 0.1 & $d_e$: object to end-effector distance, $d_g$: object to goal distance, \ \ \ \ \ \ \ \ $d_e \ tol$ params - bounds: (0, 0.05), margin: 0.3, value at margin: 0.1, $d_g \ tol$ params - bounds: (0, 0.15), margin: 0.1, value at margin:~0.1
            \\ \midrule
		\multicolumn{4}{c}{RC Car}                                                                                                   \\ \midrule
            Parking                 &      $ tol(d) +
                                            0.5 * tol(\theta) $  & 0.05 &  $d$: distance to goal position, $\theta$: angular deviation from goal, \ \ \ \ \ \ \ \ \ \ \ \ \ \ \ \ $tol$ params - bounds: (0, 0.1), margin: 0.5, value at margin: 0.2                                                   
            \\ \bottomrule
	\end{tabular}
\end{table*}

\section{Additional Experiment Results}

\subsection{SAC variant of PACOH-RL} \label{sec:sacPacohRL}
We compare the performance of the SAC variant of PACOH-RL with the MPC version used in \Cref{sec:sim_experiments} in \Cref{fig:sim_sac}. In particular, we report analogous PACOH-RL (SAC) results on the same 4 simulated environments and also compare the non-meta learning counterpart of PACOH-RL (SAC), referred to as MBPO (SAC) \cite{janner2021trust}.

We observe that PACOH-RL (SAC) performs comparably to PACOH-RL (iCEM) in almost all environments, both in terms of sample efficiency and asymptotic performance. PACOH-RL (SAC) usually achieves a lower reward in the first few episodes compared to the MPC version. This can be attributed to the sub-optimality of the SAC policy upon initialization, whereas shooting-based MPC methods can work optimally without the need for (re-)training the policy. The only exception is the Half-Cheetah environment, where the SAC versions achieve considerably better results after 25 episodes. We hypothesize that this is due to the comparably high-dimensional state space of the Half-Cheetath environment. Specifically, solving the MPC optimization problem in \Cref{eq:mpc_prob} becomes much harder for larger state-action spaces and sampling-based methods such as iCEM are particularly susceptible to this curse of dimensionality. 

Most importantly, PACOH-RL (SAC) achieves considerably higher rewards than MBPO (SAC). Again, this demonstrates the effectiveness of our meta-learned priors in improving the sample efficiency of model-based RL.

\subsection{Model-free RL baselines}
In this section, we compare the performance of our model-based meta RL algorithms (\algname and its SAC variant \algname (SAC)) to two model-free RL algorithms: PPO~\cite{ppo} and SAC~\cite{sac}, and a model-free meta-RL algorithm: RL${}^2$~\cite{duan2016rl}. We report the training results on the same 4 simulated environments as \Cref{sec:sim_experiments} in \Cref{fig:sim_mf}. We train PPO and SAC directly on the evaluation tasks, while for RL${}^2$, we first train the policy on similar amounts of meta-training data as \algname.

\begin{figure*}[!t]
    \centering
    \includegraphics[width=\textwidth]{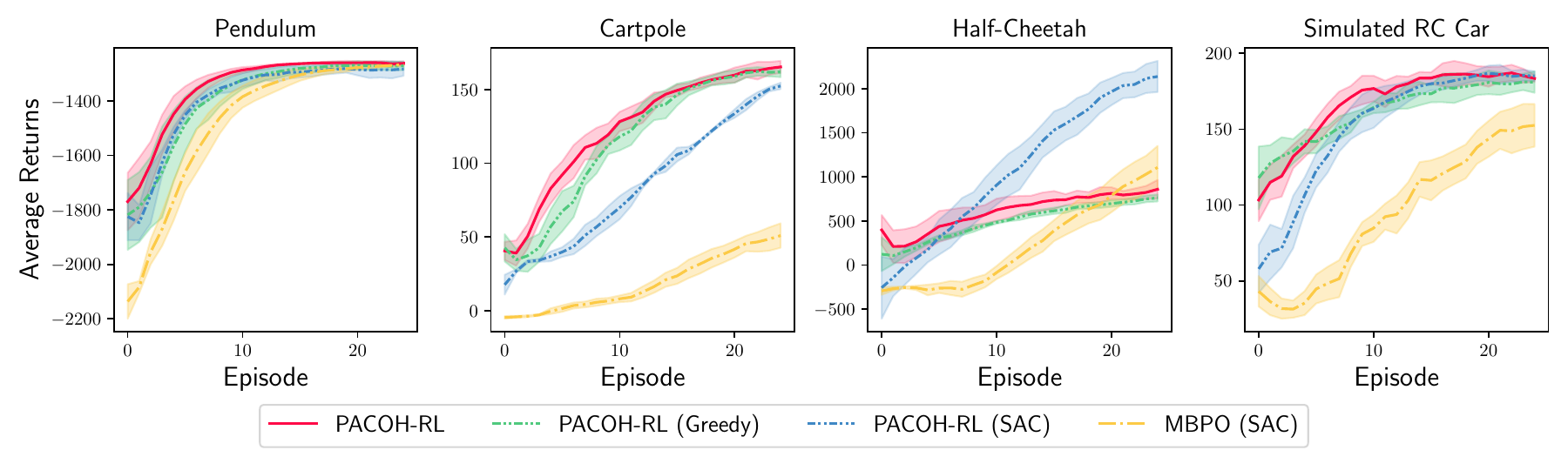}
    \caption{Returns on evaluation tasks averaged over five different seeds. We compare \algname and \algname (greedy) to our SAC-policy-based algorithm \algname (SAC). We also compare \algname (SAC) to its non-meta learning counterpart MBPO (SAC)~\cite{janner2021trust}. For all the environments, \algname (SAC) achieves similar performance to \algname and systematically outperforms the non-meta learning baseline in terms of sample efficiency and average return.}
    \label{fig:sim_sac}
\end{figure*}

\begin{figure*}[!t]
    \centering
    \includegraphics[width=\textwidth]{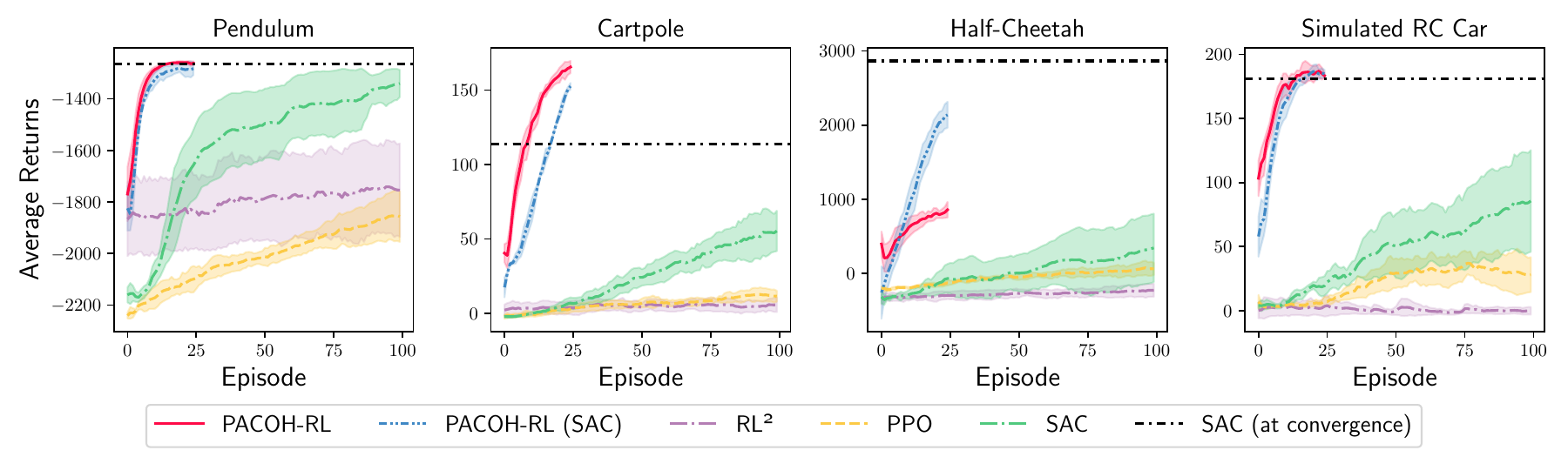}
    \caption{Returns on evaluation tasks averaged over five different seeds. We compare \algname and \algname (SAC) to model-free RL (PPO~\cite{ppo} and SAC~\cite{sac}) and model-free meta-RL (RL${}^2$~\cite{duan2016rl}) algorithms. Our algorithm achieves an order of magnitude higher sample efficiency than model-free RL algorithms in all environments.}
    \label{fig:sim_mf}
\end{figure*}

\begin{figure*}[th]
    \centering
    \includegraphics[width=\textwidth]{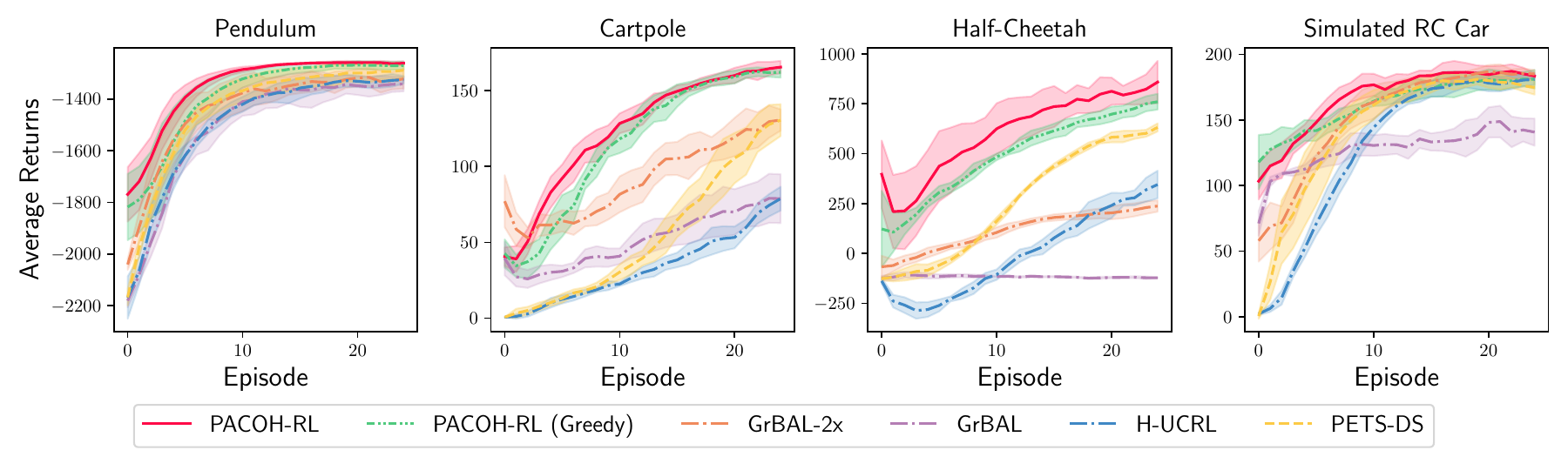}
    \caption{Returns on evaluation tasks averaged over five seeds. We compare \algname to its greedy counterpart, \algname (greedy), GrBAL \cite{nagabandi2019learning}, GrBAL-2x, H-UCRL \citep{curi2020efficient}, and PETS-DS \citep{chua2018deep}. For all the environments, \algname systematically outperforms the baselines in terms of sample efficiency and average return.}
    \label{fig:sim_normal_test_grbal2x}
\end{figure*}

\begin{table*}[t]
\renewcommand{\arraystretch}{1.15}
\caption{Compute times of PACOH-RL, GrBAL, and H-UCRL.
 The compute times are averaged over different runs and we provide the corresponding standard deviations. 
 The total compute time is split between three phases (\emph{i}) meta-learning, (\emph{ii}) model-training, and (\emph{iii}) planning.  The values for model training and planning are reported over the entire experiment, i.e. for a rollout of 25 episodes.
 We report both the absolute and relative compute time.
  }
  \centering
  \begin{tabular}{p{0.175\textwidth}>{\centering\arraybackslash}m{0.135\textwidth}>{\centering\arraybackslash}m{0.132\textwidth}>{\centering\arraybackslash}m{0.132\textwidth}>{\centering\arraybackslash}m{0.132\textwidth}>{\centering\arraybackslash}m{0.138\textwidth}}
    \toprule
    \multirow{3}{*}{\textbf{Agent}} & \multirow{3}{*}{\textbf{Phase}} & \multicolumn{4}{c}{\textbf{Average Computational Time}} \\
    \cmidrule{3-6}
           & & \textbf{Pendulum} & \textbf{Cartpole} & \textbf{Half-Cheetah} & \textbf{Simulated RC Car} \\
    \midrule
    \multirow{4}{*}{\textbf{PACOH-RL}}     & Meta-training   & 1279\si{\second}  $\pm$ 153\si{\second} \ (5.58\%) & 1568\si{\second} $\pm$ 182\si{\second} \ (9.33\%)   & 16601\si{\second}  $\pm$ 2035\si{\second} \ (23.85\%) & 2087\si{\second} $\pm$ 194\si{\second} \ (5.72\%) \\
    \cmidrule{2-6}
                                           & Model-training  & 1243\si{\second}  $\pm$ 84\si{\second}   \ (5.42\%) & 1642\si{\second} $\pm$ 116\si{\second} \ (9.77\%)   & 2167\si{\second} $\pm$ 174\si{\second} \ (3.11\%)   & 1831\si{\second} $\pm$ 148\si{\second} \ (5.02\%) \\
    \cmidrule{2-6}
                                           & Planning        & 20417\si{\second} $\pm$ 2455\si{\second} \ (89.00\%)& 13605\si{\second} $\pm$ 1637\si{\second} \ (80.90\%) & 50846\si{\second} $\pm$ 3833\si{\second} \ (73.04\%) & 32560\si{\second} $\pm$ 3149\si{\second} \ (89.25\%) \\
    \cmidrule{2-6}
                                           & \textbf{Total}  & 22939\si{\second} $\pm$ 2213\si{\second} \ (100 \%)         & 16815\si{\second} $\pm$ 1646\si{\second} \ (100 \%)           & 69614\si{\second} $\pm$ 5538\si{\second}  \ (100 \%)       & 36478\si{\second} $\pm$ 3402\si{\second}       \ (100 \%)   \\
    \midrule
    \multirow{4}{*}{\textbf{GrBAL}}        & Meta-training   & 9846\si{\second} $\pm$ 84\si{\second} \ (32.05\%)  & 6183\si{\second} $\pm$ 112\si{\second}  \ (33.32\%)  & 29084\si{\second} $\pm$ 186\si{\second} \ (44.58\%)  & 5636\si{\second} $\pm$ 66\si{\second} \ (20.92\%)\\
    \cmidrule{2-6}
                                           & Model-training  & 714\si{\second} $\pm$ 102\si{\second}  \ (2.32\%) & 691\si{\second} $\pm$ 95\si{\second} \ (3.11\%)   & 993\si{\second} $\pm$ 124\si{\second} \ (1.52\%)   & 2433\si{\second} $\pm$ 232\si{\second} \ (9.03\%)  \\
    \cmidrule{2-6}
                                           & Planning        & 20156\si{\second} $\pm$ 1539\si{\second} \ (65.62\%)& 11684\si{\second}  $\pm$ 1291\si{\second} \ (73.04\%) & 35162\si{\second} $\pm$ 2373\si{\second} \ (53.90\%) & 18870\si{\second} $\pm$ 2056\si{\second} \ (70.5\%) \\
    \cmidrule{2-6}
                                           & \textbf{Total}  & 30716\si{\second} $\pm$ 1401\si{\second}  \ (100 \%)         & 18558\si{\second} $\pm$ 994\si{\second} \ (100 \%)            & 65239\si{\second} $\pm$ 2445\si{\second}   \ (100 \%)       & 26939\si{\second} $\pm$ 2121\si{\second} \ (100 \%) \\
    \midrule
    \multirow{4}{*}{\textbf{H-UCRL}}       & Meta-training   & -  \ \ \ \ \ \ \ \ \ \ \ \ \ \ \ \ \ \ \ \ \ \ \ \ \ \ \ \ \ \ \ \ &  -  \ \ \ \ \ \ \ \ \ \ \ \ \ \ \ \ \ \ \ \ \ \ \ \ \ \ \ \ \ \ \ \  &  -  \ \ \ \ \ \ \ \ \ \ \ \ \ \ \ \ \ \ \ \ \ \ \ \ \ \ \ \ \ \ \ \  &  -  \ \ \ \ \ \ \ \ \ \ \ \ \ \ \ \ \ \ \ \ \ \ \ \ \ \ \ \ \ \ \ \  \\
    \cmidrule{2-6}
                                           & Model-training  & 1286\si{\second}  $\pm$ 67\si{\second}  \ (6.39\%) & 1502\si{\second}  $\pm$ 125\si{\second} \ (10.63\%)  & 2184\si{\second} $\pm$ 153\si{\second}  \ (4.40\%)  & 1991\si{\second} $\pm$ 150\si{\second} \ (5.26\%) \\
    \cmidrule{2-6}
                                           & Planning        & 18844\si{\second} $\pm$ 1847\si{\second} \ (93.61\%)& 12624\si{\second} $\pm$ 1438\si{\second} \ (89.37\%) & 47420\si{\second}  $\pm$ 3212\si{\second} \ (95.60\%) & 35827\si{\second} $\pm$ 3472\si{\second} \ (94.74\%) \\
    \cmidrule{2-6}
                                           & \textbf{Total}  & 20130\si{\second} $\pm$ 1740\si{\second} \  (100 \%)         & 14126\si{\second} $\pm$ 1306\si{\second} \ (100 \%)           & 49604\si{\second} $\pm$ 2762\si{\second}  \ (100 \%)       & 37818\si{\second} $\pm$ 3085\si{\second} \ (100 \%)\\
    \bottomrule
  \end{tabular}
  \label{tab:compute_times}
\end{table*}

\looseness -1 We observe that \algname and \algname (SAC) learn much faster than the model-free methods and can quickly obtain high rewards. Our algorithms achieve an order of magnitude higher sample efficiency than model-free methods while maintaining the asymptotic performance of the model-free RL methods. The only exception is the Half-Cheetah environment where \algname fails to achieve similar performance to SAC and \algname (SAC). As discussed in \Cref{sec:sacPacohRL}, we hypothesize this is due to the high dimensionality of the problem, which is challenging for a sampling-based optimizer such as iCEM to solve. For the RL$^2$ agent, a limited number of meta-training tasks leads to negative transfer, and the policy fails to perform on evaluation tasks, often performing poorer than non-meta RL methods.

\subsection{GrBAL with more tasks}
We ran experiments where we gave GrBAL twice the number of meta-training tasks (GrBAL-2x) as shown in \cref{fig:sim_normal_test_grbal2x}. As we can observe, GrBAL performs better when provided with more meta-training data. This indicates that GrBAL generally works but struggles in realistic robotic settings where we only have a few tasks for meta-learning. Conversely, PACOH-RL performs significantly better in the low-data regime is due to its principled treatment of uncertainty and meta-level regularization.

\subsection{Computational Complexity}
We perform experiments to quantify the additional computational complexity required for meta-learning the dynamics prior. The total runtimes for different agents and environments are reported in \cref{tab:compute_times} along with the relative compute times required for individual phases. All experiments are performed on two cores of a 2.6~GHz~AMD~EPYC~7H12~CPU. We average the compute times over different runs and provide the corresponding standard deviations. The values for model training and planning are reported over the entire experiment, i.e. for a rollout of 25 episodes. 

As expected, the meta-learning of dynamics priors results in additional computational complexity compared to standard model-based RL. However, the computational burden of meta-learning is typically outweighed by the magnitude of more compute-intensive policy search or MPC-based planning. Crucially, meta-learning the prior only happens once, whereas the dynamics model and SAC policy have to be re-trained after every episode. Since, with the meta-learned priors, we typically need much fewer episodes to reach a comparable performance, PACOH-RL typically results in significant net savings in compute time. \Cref{tab:compute_times} validates our argument above that planning requires the majority of compute time and meta-learning the prior is relatively cheap in comparison.

}{}

\clearpage
\addtolength{\textheight}{-6cm}   %

\newpage

\end{document}